%% file: full_manuscript.tex
\documentclass[runningheads]{llncs}

\usepackage{eccv}

\usepackage{eccvabbrv}

\usepackage{graphicx}

\usepackage{array}
\usepackage{multirow}

\usepackage{tcolorbox}
\tcbuselibrary{listings, breakable, skins}

\usepackage{array}      % Required for table formatting
\usepackage{booktabs}   % Required for toprule, midrule, bottomrule
\usepackage{longtable}  % <--- UNCOMMENTED THIS so the table works
\usepackage{ragged2e}   % Required for text alignment
\usepackage[accsupp]{axessibility}  % Improves PDF readability for those with disabilities.

\usepackage{algorithm}
\usepackage{algpseudocode}
\usepackage{xr-hyper}

\usepackage{hyperref}

\usepackage{orcidlink}

\newcommand{\eg}{\textit{e.g.\ }}

\input{preamble}

\begin{document}

%\title{DynaEdit: Training-Free Structure-Evolving Video Editing With Interaction Dynamics} 
% \title{DynaEdit: Training-Free Editing of Motion, Appearance, and Interactions in Videos} 
% \title{Beyond structure preservation: Training-Free Editing of Motion, Appearance, and Interactions in Videos} 

\title{Versatile Editing of Video Content, Actions, and Dynamics without Training}

% \titlerunning{Training-Free Structure-Unbound Text-Based Video Editing With Flow Models}

\author{Vladimir Kulikov\thanks{Work done during an internship at Google DeepMind.}\inst{1, 2} %\orcidlink{0009-0001-5963-2083} 
\and
Roni Paiss\inst{1} \and
Andrey Voynov\inst{1} \and
Inbar Mosseri\inst{1} \and
Tali Dekel\inst{1,3} \and
Tomer Michaeli\inst{2}
}

\authorrunning{V.~Kulikov et al.}

\institute{Google DeepMind
 \and
Technion -- Israel Institute of Technology  \and
The Weizmann Institute of Science
}

\maketitle

\input{sections/main_paper/all_sections}

% \clearpage\mbox{}Page \thepage\ of the manuscript.
% \clearpage\mbox{}Page \thepage\ of the manuscript.
% \clearpage\mbox{}Page \thepage\ of the manuscript.
% \clearpage\mbox{}Page \thepage\ of the manuscript.
% \clearpage\mbox{}Page \thepage\ of the manuscript. This is the last page.
% Now we have reached the maximum length of an ECCV \ECCVyear{} submission (excluding references).
% References should start immediately after the main text, but can continue past p.\ 14 if needed.
\clearpage  % TODO REVIEW/FINAL: This \clearpage needs to be removed from both review and camera-ready versions.

\bibliographystyle{splncs04}
\bibliography{main}

% add sm here for now
% \beginsupplement
% \clearpage
\newpage
\appendix

\author{}
\institute{}

\authorrunning{V.~Kulikov et al.}

\input{appendix}

\end{document}

%% file: preamble.tex
\newcommand{\ours}[1]{DynaEdit}

\makeatletter
\newcommand*{\addFileDependency}[1]{% argument=file name and extension
  \typeout{(#1)}
  \@addtofilelist{#1}
  \IfFileExists{#1}{\typeout{File #1 O.K.}}{\typeout{No file #1.}}
}
\makeatother

%% file: sections/main_paper/all_sections.tex
\begin{figure}[t]
\centering
%\captionsetup{type=figure}
\includegraphics[width=\textwidth]{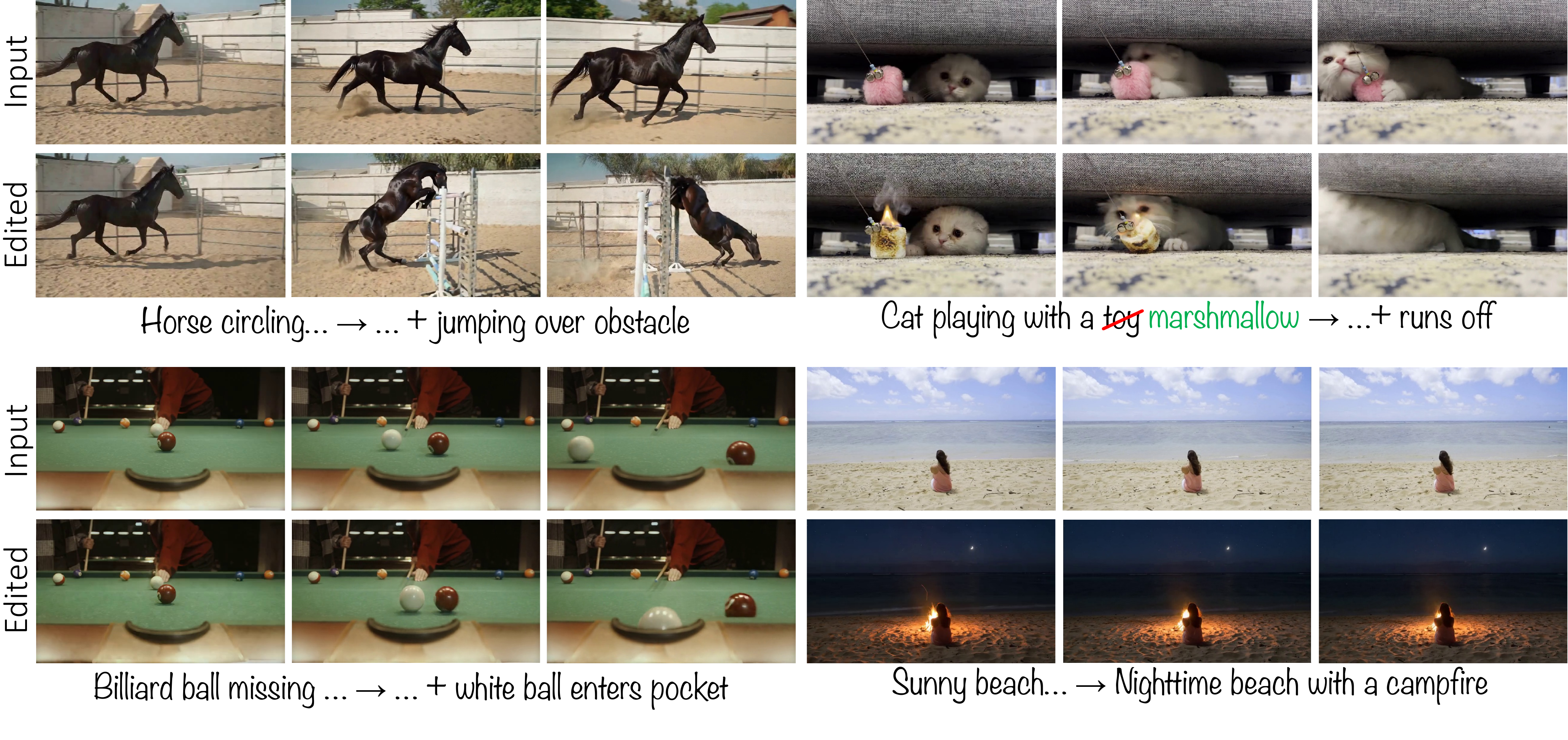}
\caption{\textbf{Training-free versatile 
editing of actions and dynamics in videos.}
We present \ours{}, a training-free flow-based method for video editing, which is the first to enable manipulation of dynamics and contents in videos using textual descriptions. \ours{} supports the modification of actions and the insertion of objects that interact with the scene (\eg causing a horse to jump due to a newly inserted obstacle, a cat to run off due to a toy edited to become a  burning marshmallow, or a billiard ball to enter the pocket). It also allows global and stylistic modifications, like changing daytime to nighttime, all while avoiding unnecessary changes to the video (see SM for the videos).}%\tali{something is off with the caption -- there is body text in-between the figure and the caption.}}
\label{fig:teaser}
\end{figure}%

\begin{abstract} 
Controlled video generation has seen drastic improvements in recent years. 
However, editing actions and dynamic events, or inserting contents that should affect the behaviors of other objects in real-world videos, remains a major challenge. Existing trained models struggle with complex edits, likely due to the difficulty of collecting relevant training data. Similarly, existing training-free methods are inherently restricted to structure- and motion-preserving edits and do not support modification of motion or interactions. 
Here, we introduce \ours{}, a training-free editing method that unlocks versatile video editing capabilities with pretrained text-to-video flow models. 
Our method relies on the recently introduced inversion-free approach, which does not intervene in the model internals, and is thus model-agnostic. We show that naively attempting to adapt this approach to general unconstrained editing results in  severe low-frequency misalignment and high-frequency jitter. We explain the sources for these phenomena and introduce novel mechanisms for overcoming them. 
Through extensive experiments, we show that 
\ours{} achieves state-of-the-art results on complex text-based video editing tasks, including modifying actions, inserting objects that interact with the scene, and introducing global effects (see \href{https://dynaedit.github.io/}{website}).

\end{abstract}

\section{Introduction}
\label{intro}

Generative video models have advanced to a point where synthesized content is increasingly indistinguishable from reality in its adherence to physics, causality, and complex dynamics \cite{bartal2024lumierespacetimediffusionmodel, blattmann2023alignlatentshighresolutionvideo, HaCohen2024LTXVideo, kong2024hunyuanvideo, wan2025, yang2025cogvideoxtexttovideodiffusionmodels,hacohen2026ltx2efficientjointaudiovisual,GoogleVeo2024}.
Modern text-to-video models are now often regarded as ``world models'' -- foundation models that possess an inherent understanding of our physical and dynamic world \cite{motamed2025generativevideomodelsunderstand, wiedemer2025videomodelszeroshotlearners}.  Given this progress, a natural question arises -- can we tap into the immense knowledge of these models to alter a real-world video rather than generating one from scratch? For example, can we change the actions and movements of a subject, insert or swap an existing object to facilitate meaningful interaction with the scene, or create global effects that integrate naturally with the world? 

Despite the remarkable progress in video editing \cite{wu2023tuneavideooneshottuningimage,qi2023fatezerofusingattentionszeroshot,yang2023rerendervideozeroshottextguided,geyer2023tokenflowconsistentdiffusionfeatures,cohen2024sliceditzeroshotvideoediting, kara2023raverandomizednoiseshuffling,li2025flowdirector0,wang2025tamingrectifiedflowinversion,cong2024flattenopticalflowguidedattention,ceylan2023pix2videovideoeditingusing,liu2023videop2pvideoeditingcrossattention,wang2025videodirectorprecisevideoediting,singer2024videoeditingfactorizeddiffusion}, the task of non-rigid, dynamic manipulation in real-world videos remains an open challenge. This stems from a fundamental tension in the editing objective: the model must possess enough flexibility to fundamentally alter motion or object interactions, yet simultaneously remain strictly faithful to the original objects' identities and environmental context. A data-driven approach to this problem is hindered by the difficulty of obtaining high-quality training data. Specifically, non-rigid editing requires precisely paired source-target example videos that demonstrate the same scene under different physical outcomes, data that is exceptionally difficult to collect or simulate at scale. 
Currently, RunwayML's Gen-4 Aleph \cite{RunwayAleph2025} is the only publicly available trained model that provides a general prompt-based framework for manipulation of video. While constituting a significant advancement, this model still struggles with complex non-rigid action-altering edit requirements.
Several works proposed training-free editing methods that harness a pre-trained text-to-video model \cite{cong2024flattenopticalflowguidedattention, geyer2023tokenflowconsistentdiffusionfeatures, cohen2024sliceditzeroshotvideoediting,ouyang2024i2vedit, kim2025flowaligntrajectoryregularizedinversionfreeflowbased,li2025flowdirector0,ku2024anyv2v,kulikov2025flowedit}. Yet, these methods are constrained to structurally aligned transformations, or to layer-like object insertion, where the inserted object can one-sidedly react to the rest of the content in the video, but cannot affect it.

In this paper, we introduce \ours{}, a training-free method for in-the-wild unconstrained video editing. Given an input source video and a target text prompt that describes the edit, our method steers the generation process of a pre-trained text-to-video flow model towards the desired solution -- altering the scene's dynamics, while preserving the properties of the original video that should not be affected by the edit. As shown in Fig.~\ref{fig:teaser}, \ours{} supports modification of dynamic events, like causing a horse to jump over a newly inserted obstacle, a billiard ball to enter the pocket, or a cat to run off due to interaction with a toy that was edited to become a burning marshmallow. It also allows global modifications, like changing a sunny scene into a nighttime setting.

\ours{} relies on the recently introduced inversion-free approach \cite{kulikov2025flowedit}. 
We show that the naive adaptation of this approach to support significant spatio-temporal modifications leads to severe \textit{low-frequency misalignment} with the source video and \textit{high-frequency jitter}. We explain why these phenomena arise and introduce two novel components for mitigating them: a Similarity Guided Aggregation (SGA) mechanism and an Annealed Noise Correlation (ANC) schedule. Extensive evaluations demonstrate that \ours{} not only outperforms all existing training-free methods but effectively closes the performance gap with the proprietary trained Aleph model on a wide variety of complex editing tasks.

\section{Related Work}
\label{related}

Text-to-video generative models~\cite{bartal2024lumierespacetimediffusionmodel, blattmann2023alignlatentshighresolutionvideo, HaCohen2024LTXVideo, kong2024hunyuanvideo, wan2025, yang2025cogvideoxtexttovideodiffusionmodels,hacohen2026ltx2efficientjointaudiovisual, GoogleVeo2024,openai2025sora2} have seen tremendous recent progress, with the most advanced open-source models~\cite{HaCohen2024LTXVideo,wan2025,kong2024hunyuanvideo} relying on the flow matching framework \cite{lipman2023flowmatchinggenerativemodeling, liu2022flowstraightfastlearning}. This progress has given rise to numerous video editing methods. Many methods target specific types of edits, such as motion transfer \cite{pondaven2025videomotiontransferdiffusion, meral2024motionflowattentiondrivenmotiontransfer,yatim2023spacetimediffusionfeatureszeroshot,jiang2025vaceallinonevideocreation}, effect transfer \cite{jones2026tuningfreevisualeffecttransfer}, object insertion \cite{tu2025videoanydoorhighfidelityvideoobject, tewel2024addittrainingfreeobjectinsertion, yatim2025dynvfxaugmentingrealvideos,bai2024scenephotorealisticvideoobject}, optical-flow or keypoint-controlled motion editing \cite{burgert2025motionv2veditingmotionvideo, burgert2025gowiththeflowmotioncontrollablevideodiffusion}, re-angling~\cite{Zhang2024ReCaptureGV,Wu2024CAT4DCA} or style transfer \cite{ye2024stylemasterstylizevideoartistic, mehraban2025pickstylevideotovideostyletransfer}.
Here, we focus on general-purpose video editing using only text. Existing methods in this category focus on the sub-task of structurally-aligned editing \cite{kim2025flowaligntrajectoryregularizedinversionfreeflowbased, li2025flowdirector0, ouyang2024i2vedit, ku2024anyv2v}, leaving general editing an open challenge.

\vspace{-0.2cm}
\paragraph{Training-based video editing.} 
Training a model for general video editing requires non-trivial data collection and an extensive computational budget. %, current models that are trained for in-the-wild editing are exclusively owned by companies and are not open sourced to the community.
Some methods propose lightweight inference-time training \cite{gao2025lora,polaczek2025incontextsyncloraportraitvideo}.
The only trained model that currently supports in-the-wild video editing is RunwayML's Gen-4 Aleph~\cite{RunwayAleph2025}, which is not open source. This model was the first to allow text-based editing of actions and dynamics, however it still struggles to perform complex manipulations, attesting to the difficulty of the task.

\vspace{-0.2cm}
\paragraph{Training-free video editing.} 
Opting for open source video editing solutions, several works proposed training-free methods which utilize pre-trained video flow models. These works can be broadly categorized into inversion-based and inversion-free approaches. 
Inversion-based methods, such as~\cite{wang2025tamingrectifiedflowinversion, yatim2025dynvfxaugmentingrealvideos,meral2024motionflowattentiondrivenmotiontransfer}, start by finding a noise initialization that reconstructs the input video when conditioned on a source prompt describing that video. They then use that noise for sampling a new video by conditioning on a target prompt that describes the desired edit. This approach by itself often leads to poor results%mainly because of limited reconstruction quality and suboptimal coupling between the source and target distributions
~\cite{hubermanspiegelglas2024editfriendlyddpmnoise} and therefore many works proposed additional model-specific interventions. For example, DynVFX~\cite{yatim2025dynvfxaugmentingrealvideos} employ attention-based manipulations during the inversion-and-sampling process to perform object insertion. This method can incorporate objects in a natural harmonious manner, however the inserted objects cannot dynamically interact with the surrounding scene or alter the video's outcomes.
Inversion-free approaches~\cite{kulikov2025flowedit,li2025flowdirector0,kim2025flowaligntrajectoryregularizedinversionfreeflowbased} traverse a noise-free path between the source and target domains, without relying on inversion. 
FlowEdit~\cite{kulikov2025flowedit} first proposed and implemented this paradigm for flow-based image editing. FlowAlign~\cite{kim2025flowaligntrajectoryregularizedinversionfreeflowbased} introduced an improved variant of this approach and exemplified its effectiveness in the video domain as well. FlowDirector~\cite{li2025flowdirector0} proposed an ad-hoc solution for swapping objects in videos by leveraging an attention-based mask construction to constrain the edits to desired regions.
While these approaches can achieve better quality than inversion-based methods, both approaches are constrained to strong structure-preserving edits, with limited ability to change the coarsest features of the source video. 
In this work, we build upon the inversion-free editing approach, but make key adaptations to allow it to support structurally unrestricted editing.

\section{Preliminaries}
\label{prelim}

We use upper-case and lower-case letters to denote random variables and their realizations (samples from the corresponding distribution), respectively.

\subsection{Rectified Flow Models}
Flow models learn a velocity field $V$, parameterized by a neural network, with which they generate samples by solving the ODE
\begin{equation}\label{eq:flow_rewritten_2}
    dZ_t=V(Z_t, t) \,dt
\end{equation}
over $t\in[0,1]$. The core objective is to have the ODE transport from a simple prior at $t=1$, typically $\mathcal{N}(0, \boldsymbol{I})$, to the data distribution at $t=0$. Sampling thus involves initializing the ODE at $t=1$ with a sample of Gaussian noise and numerically solving it in reverse down to $t=0$. In practice, the integration is performed over $N$ discrete steps $\{t_i\}^N_{i=0}$. Rectified flows \cite{albergo2025stochasticinterpolantsunifyingframework, lipman2023flowmatchinggenerativemodeling, liu2022flowstraightfastlearning} represent a specific class of these models, where $Z_t$ is distributed like
\begin{equation}\label{eq:linear_interpolant_1}
    X_t = (1-t) X_0 + t X_1,
\end{equation}
where $X_0,X_1$ are statistically independent. This choice leads to low path curvatures and thus enables sampling with a small number of discretization steps.

\begin{figure}[t]
    \centering
    \includegraphics[width=\textwidth]{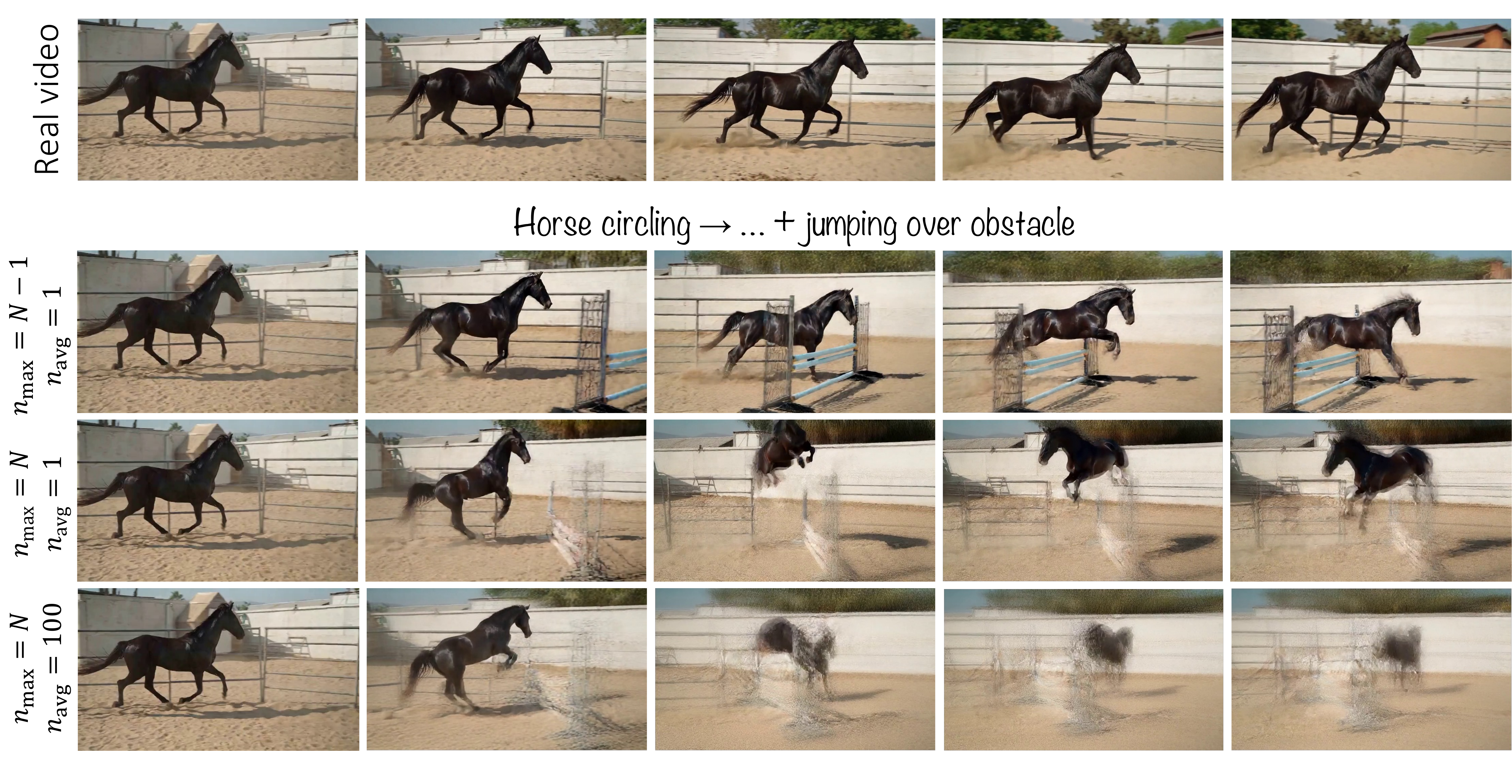}
    \caption{\textbf{Limitations of state-of-the-art inversion-free editing methods.} Existing inversion-free methods \cite{kulikov2025flowedit,li2025flowdirector0,kim2025flowaligntrajectoryregularizedinversionfreeflowbased} suffer from a tradeoff between edit expressivity and visual quality, illustrated here with FlowEdit \cite{kulikov2025flowedit} using an I2V model. When starting the generation at timestep $n_{\text{max}}=N-1$, the method struggles to modify motion (second row). Using $n_{\text{max}}=N$ allows making more significant spatio-temporal modifications and thus to better adhere to the prompt, but results in severe jitter artifacts and illogical motions (third row). Attempting to reduce artifacts by averaging over $n_{\text{avg}}=100$ edit directions in each step leads to blur (fourth row).
    }
    \label{fig:ours_vs_fe}
\end{figure}

Image-to-video (I2V) flow models employ a velocity field $V(x_t, t, c, f)$ that is conditioned on a text prompt $c$ and an image $f$ depicting the first frame. Such models are trained on triplets of text, first frame, and video data, $\{c, f, x_0\}$, and thus enable sampling from the conditional distribution of $X_0$ given $C, F$. 
Throughout this work we use an I2V model, which is beneficial for our task of general video editing. Specifically, when the edited video is required to lose spatio-temporal alignment with the source video, the first frame conditioning helps in maintaining scene, object, and color-palette consistency (see App.~\ref{sec:sm_i2v_cond}).

\subsection{Inversion-Free Editing With Pretrained Flow Models}

In text-based video editing, the user provides an input video $x^{\text{src}}$, a source prompt describing it $c^{\text{src}}$, a target prompt $c^{\text{tar}}$ that describes the desired edit, and optionally an edited first frame $f^{\text{tar}}$ for added conditioning. Several approaches exist for inversion-free editing \cite{kulikov2025flowedit,kim2025flowaligntrajectoryregularizedinversionfreeflowbased, li2025flowdirector0}. Here we focus on FlowEdit \cite{kulikov2025flowedit}.

The idea in this approach is to construct an ODE that directly transforms the source video into an edited video, such that all intermediate videos along the path are noise-free. 
To simplify notations we denote the source- and target-conditioned velocities by $V^{\text{src}}(x_t,t)=V(x_t, t, c^\text{src}, f^{\text{src}})$ and $V^{\text{tar}}(x_t,t)=V(x_t, t, c^\text{tar}, f^{\text{tar}})$, respectively. In FlowEdit, the noise-free path is traced by the ODE
\begin{equation}
\label{eq:FE_ODE}
    dZ^{\text{edit}}_t= \mathbb{E}\left[V^{\Delta}_{t}(Z^{\text{src}}_{t},Z^{\text{tar}}_{t}) \right]\,dt ,
\end{equation}
where $V^{\Delta}_{t}(Z^{\text{src}}_{t},Z^{\text{tar}}_{t})=V^{\text{tar}}(Z^{\text{tar}}_{t},t)-V^{\text{src}}(Z^{\text{src}}_{t},t)$. Here,  $Z^{\text{src}}_{t}=(1-t)x^{\text{src}}+tW_t$ is a noisy version of the source video obtained with $W_t\sim\mathcal{N}(0,I)$, and $Z^{\text{tar}}_{t}=Z^{\text{edit}}_t+Z^{\text{src}}_{t}-x^{\text{src}}$ is a noisy version of the target video being constructed. The expectation is over $W_t$. The ODE is initialized at $t=1$ with the source video $Z^{\text{edit}}_1=x^{\text{src}}$ and solved backwards down to $t=0$ to obtain an edited video $Z^{\text{edit}}_0$.

\begin{algorithm}[tb]
   \caption{Inversion-free editing (FlowEdit)}
   \label{alg:flowedit}
\begin{algorithmic}[1]
   \State {\bfseries Input:} 
   real video $x^{\text{src}}$, source prompt $c^\text{src}$, target prompt $c^\text{tar}$ %$,\big\{t_i\big\}^N_{i=0}, n_{\text{max}}, n_{\text{avg}}$ 
   \State {\bfseries Output:} edited video $z^{\text{edit}}$ % $X^{\text{tar}}$
   \State {\bfseries Initialize:} $z^{\text{edit}}=x^{\text{src}}$
   
   \For{$i=n_{\text{max}}$ {\bfseries to} $1$}
   \State $\{w_j\}^{n_\text{avg}}_{j=1} \sim \mathcal{N}(0,I)$
   \textcolor{blue}{// Sample random noise vectors}
   \State $z^{\text{src}}_{j} \leftarrow (1-t_i)x^{\text{src}} + t_iw_j$
   \textcolor{blue}{// Construct noisy source samples}
   \State $z^{\text{tar}}_{j} \leftarrow z^{\text{edit}} +z^{\text{src}}_{j}- x^{\text{src}}$
   \textcolor{blue}{// Construct noisy target samples}
   \State $V^{\Delta}_{j} \leftarrow V(z^{\text{tar}}_{j},t_i,c^\text{tar})-V(z^{\text{src}}_{j},t_i,c^\text{src})$   
   \textcolor{blue}{//Calculate velocity differences}
   \State $\bar{V}^{\Delta} \leftarrow \frac{1}{n_\text{avg}}\sum^{n_\text{avg}}_{j=1}V^{\Delta}_{j}$
   \textcolor{blue}{// Average directions}
   \State $z^{\text{edit}} \leftarrow z^{\text{edit}} + (t_{i-1}-t_{i})\bar{V}^{\Delta}$
   \textcolor{blue}{// Propagate ODE}
   \EndFor
   \State {\bfseries Return:} $z^{\text{edit}}$% = x_{0}^{\text{tar}}$
\end{algorithmic}
\end{algorithm}

In practice, the expectation in \eqref{eq:FE_ODE} is approximated by averaging over $n_{\text{avg}}$ independent noise samples in each timestep. These samples are taken to be independent also across timesteps, a fact that turns out to play an important role, as we illustrate in Secs.~\ref{sec:roadblocks} and \ref{sec:method}. %When using a sufficiently dense time discretization $\{t_i\}_{i=0}^N$, 
The hyperparameter  $n_{\text{avg}}$ is often set to~$1$, as averaging naturally occurs also across timesteps.

To control the amount of deviation from the source video, FlowEdit can be initialized at a timestep $n_{\text{max}}\leq N$. This hyperparameter effectively controls the maximum amount of noise that is added to the source video and thus implicitly determines the coarsest spatio-temporal features that can get modified. It therefore controls the tradeoff between edit expressivity and structural adherence to the source video. 
A pseudo-code for this method is given in Alg.~\ref{alg:flowedit}, where subscripts are used to index samples within the batch rather than time.

\section{Roadblocks towards motion and interaction editing}

\label{sec:roadblocks}

Existing inversion-free methods struggle with complex edits that require significant spatio-temporal modifications. For example, while FlowEdit should in theory be able to perform arbitrary edits given a large enough $n_{\text{max}}$, it is practically impossible to select a value for $n_{\text{max}}$ that strikes a good balance between output quality, prompt adherence, and loyalty to the source video. 
This is illustrated in Fig.~\ref{fig:ours_vs_fe}, which shows FlowEdit results using $n_{\text{max}}=N-1$ and $n_{\text{max}}=N$. Here, the goal is to insert an obstacle and have the horse jump over it. As seen, setting $n_{\text{max}}=N-1$ is too restrictive for the requested edit (the horse fails to perform the requested jump).
On the other hand, setting $n_{\max}=N$ results in a video that adheres to the edit prompt, but exhibits extraneous low frequency changes (the horse's trajectory needlessly deviates from the source motion), and suffers from severe
high frequency jitter artifacts (evident by the blurry obstacle). Note that while FlowEdit's velocity averaging usually improves quality, in the case of structurally-unaligned video editing, it results in blurry edits, as seen in the last row, where $n_{\text{avg}}=100$ is used. We next analyze the causes for the low frequency misalignment and high frequency jitter that emerge when $n_{\text{max}}=N$.

\begin{figure}[t]
    \centering
    \includegraphics[width=\linewidth]{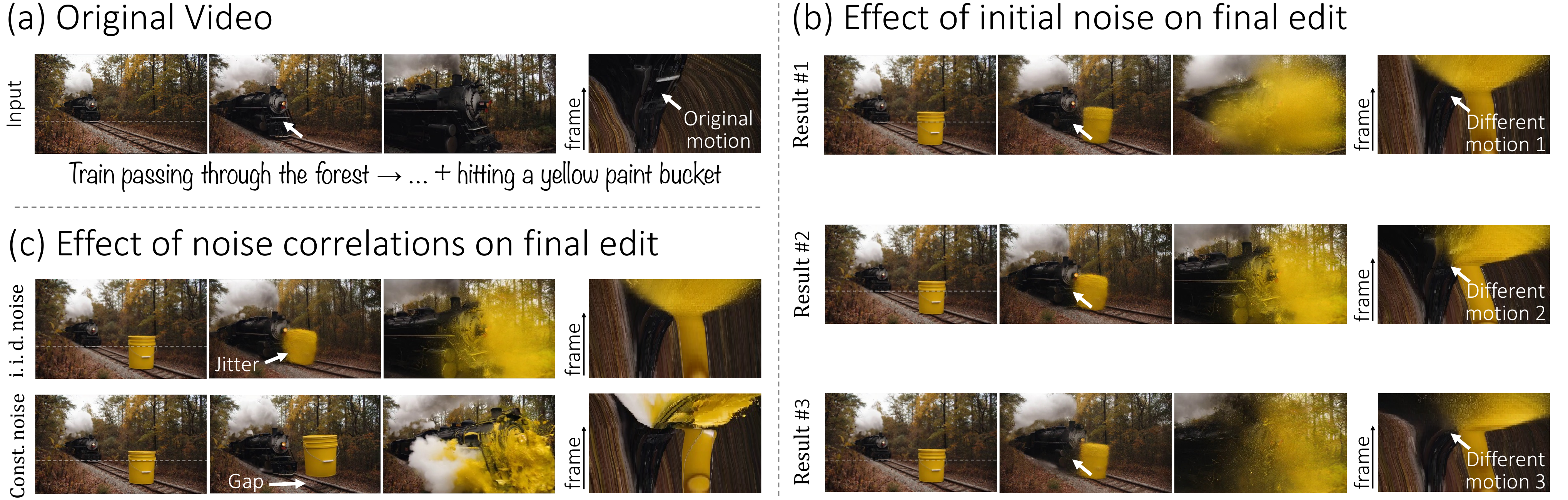}
    \caption{\textbf{Effects of noise in inversion-free editing.} (a) A source video (three frames and a spatio-temporal slice corresponding to the dashed line). (b)~Three inversion-free editing results (FlowEdit) differing only in the noise sample at timestep $t_N$. As seen, the initial noise strongly affects coarse spatio-temporal features, \eg modifying the camera motion and train position across edits, although those features are not required to change to adhere to the prompt.
    (c)~Using independent noise samples across timesteps (top) leads to high-frequency jitter, \eg the blurry bucket and paint drops. This can be alleviated by using the same noise sample for all timesteps (bottom) but at the cost of worse alignment with the source video, \eg causing the bucket to levitate.}
    \label{fig:uncertainty_combined}
\end{figure}

\vspace{-0.2cm}
\paragraph{Low frequency misalignment.}

When using $n_{\text{max}}=N$, the noisy marginals $Z^{\text{src}}_{t_N}$ and $Z^{\text{tar}}_{t_N}$ in Eqn.~\eqref{eq:FE_ODE} contain pure noise (both equal $W_{t_N}$). This means that the edit velocity $V^\Delta_{t_N}$ has no connection to the source video beyond the first frame conditioning. The effect that this has is visualized in Fig.~\ref{fig:uncertainty_combined}(a),(b). Here, the goal is to insert a bucket of paint to the train tracks and have the train collide with it. The figure shows the input video and three different edit results, each obtained with a different noise realization for the initial timestep, but the same set of noise maps for all subsequent timesteps. As seen, the resulting videos have different camera motions, train speeds, and bucket explosion times. This reveals that the initial edit step has an immense impact on the coarse spatio-temporal features of the edited video. Importantly, the resulting edits are not well-aligned with the source video, as seen by the spatio-temporal slices (\eg the curves caused by the camera motion do not align). This suggests that while using $n_{\max}=N$ is crucial for modifying coarse spatio-temporal features, the noise realizations in the initial timesteps should be carefully selected to allow maintaining adherence to the source video. We explore this in Sec.~\ref{sec:SGA}.

\vspace{-0.2cm}
\paragraph{High frequency jitter.}
When the edited video contains assets that are not spatio-temporally aligned with the source (\eg an inserted object or edited dynamics), severe high frequency jitter emerges. This is evident in the first row of Fig.~\ref{fig:uncertainty_combined}(c), where the high frequencies of the bucket and the paint drops are fuzzy. We hypothesize that this stems from the fact that the noises $\{W_{t_n}\}$ are uncorrelated across timesteps. This causes the edit velocities $V^{\Delta}_t$ to point to different directions that accumulate to the visible jitter artifacts. To test this hypothesis, the second row of Fig.~\ref{fig:uncertainty_combined}(c) shows the result obtained when using the same noise realization $W_{t_n}=W^{\text{const}}$ for all timesteps.  As can be seen, this indeed eliminates the high-frequency jitter. Unfortunately, however, it worsens the alignment with the input video's coarse features, resulting in unnatural interactions (notice the levitating bucket). This suggests that introducing some amount of correlation between the noises of different timesteps may allow to strike a good balance between visual quality and low-frequency alignment. We explore this in Sec.~\ref{sec:ANC}.

\begin{figure}[t]
    \centering
    \includegraphics[width=\textwidth]{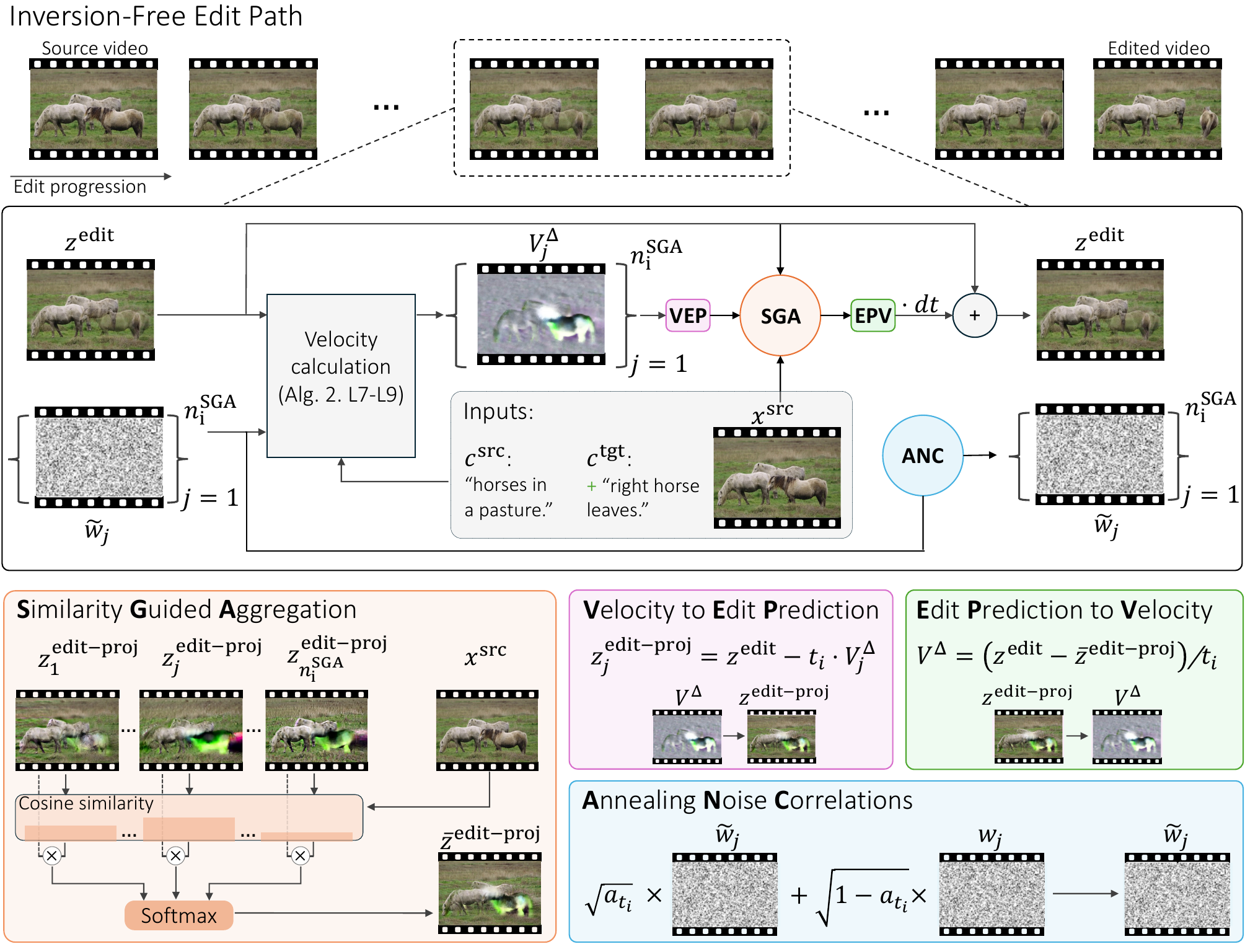}
    \caption{\textbf{\ours{}.} Our method constructs a noise free path from the source video to the edited one (top pane). The middle pane shows one step in this process, with our key contributions colored. Our SGA module (bottom left) aggregates several noise-free velocities based on the similarities between the edits they induce and the source video. Our ANC mechanism (bottom right) induces gradually increasing correlations between the noises of consecutive timesteps.
    }
    \label{fig:method_pipeline}
\end{figure}

\section{Method}
\label{sec:method}

We now present \ours{}, an inversion-free method that overcomes the limitations discussed in Sec.~\ref{sec:roadblocks}. \ours{} relies on two new components, as we detail next. The method is illustrated in Fig.~\ref{fig:method_pipeline}, and pseudo-code is provided in Alg.~\ref{alg:our_alg}.

\subsection{Similarity Guided Aggregation}
\label{sec:SGA}

In Fig.~\ref{fig:uncertainty_combined}(b), we saw that the initial edit steps facilitate the most significant changes to the low spatio-temporal frequencies, but vary significantly depending on the noise seed. To achieve edits that are better aligned to the source frequencies, we propose \textit{similarity guided aggregation} (SGA), a mechanism for soft selection of edit velocities based on their similarity to the source video. 
In each edit step~$i$, we use $n^{\text{SGA}}_i$ noise samples to obtain random edit directions $V^{\Delta}_{1},\ldots,V^{\Delta}_{n^{\text{SGA}}_i}$. We predict the final edit that would be obtained with each of them by using $\Delta t=t_0-t_i=-t_i$, namely we construct the projected edits
\begin{equation}
\label{eq:edit_proj}
z^{\text{edit-proj}}_{j}=z^{\text{edit}}-t_iV^{\Delta}_{j}.
\end{equation}
We calculate the cosine similarity between each prediction and the source video to obtain  coefficients  $s_j= \text{sim}(x^\text{src}, z^{\text{edit-proj}}_{j})$ (see App.~\ref{sec:sm_sga_sim_ablation} for the effect of other similarity metrics) 
and normalize them using softmax with temperature $\tau$. The resulting weights are used to construct the combined edit prediction as
\begin{equation}
\label{eq:SGA_V_delta}
    \bar{z}^{\text{edit}}= \sum_{j=1}^{n^{\text{SGA}}_i}s_jz^{\text{edit-proj}}_{j}.
\end{equation}
This prediction is transformed back to a velocity to obtain the edit direction
\begin{equation}
\label{eq:V_Delta_Bar}
    \bar{V}^{\Delta}= (z^{\text{edit}}-\bar{z}^{\text{edit-proj}})/t_i.
\end{equation}
The SGA module is depicted in the bottom-left pane of Fig.~\ref{fig:method_pipeline}. We find that to save computation, it is enough to use $n^{\text{SGA}}_i>1$ only for the first few timesteps (see Sec.~\ref{sec:implementation} for details). 
The softmax temperature $\tau$ controls the degree of alignment between the edited video and the source video. When $\tau$ is small Eq.~\eqref{eq:V_Delta_Bar} collapses to a hard-selection rule, retaining only the edit path that best matches the source video and thus leading to stronger alignment. We demonstrate the advantage of SGA over the simple velocity averaging of FlowEdit \cite{kulikov2025flowedit} in App.~\ref{sec:sm_ablation_SGA}.

\begin{algorithm}[tb]
   \caption{\ours{}}
   \label{alg:our_alg}
\begin{algorithmic}[1]

   \State {\bfseries Input:} 
   real video $x^{\text{src}}$, source prompt $c^\text{src}$, target prompt $c^\text{tar}$ 
   \State {\bfseries Output:} edited video $x^{\text{edit}}$
   
   \State {\bfseries Initialize:} $z^{\text{edit}}=x^{\text{src}}$, $\{\tilde{w}_j\}_{j=1}^{n^{\text{SGA}}_N}=0$
    \For{$i=N$ {\bfseries to} $1$}
           \State $\{w_j\}^{n^{\text{SGA}}_i}_{j=1} \sim \mathcal{N}(0,I)$
           \textcolor{blue}{// Sample random noise vectors}
           \State $\tilde{w}_{j} \leftarrow \sqrt{a_i}\tilde{w}_{j}+\sqrt{1-a_{t_i}}w_{j}$
           \textcolor{blue}{// Construct correlated noise with ANC (Eq.~\eqref{eq:add_noise_corr})}
           \State $z^{\text{src}}_{j} \leftarrow (1-t_i)x^{\text{src}} + t_i\tilde{w}_{j}$
           \textcolor{blue}{// Construct noisy source samples}
           \State $z^{\text{tar}}_{j} \leftarrow z^{\text{edit}} +z^{\text{src}}_{j}- x^{\text{src}}$
           \textcolor{blue}{// Construct noisy target samples}
           \State $V^{\Delta}_{j} \leftarrow V(z^{\text{tar}}_{j},t_i, c^\text{tar})-V(z^{\text{src}}_{j},t_i,c^\text{src})$
           \textcolor{blue}{// Calculate velocity differences}
           % \ENDFOR
    % \IF{$i \geq n_\text{LFR}$}
        \State $\bar{V}^{\Delta} \leftarrow \text{SGA}(\{V^{\Delta}_{j}\}^{n^{\text{SGA}}_i}_{j=1})$            \textcolor{blue}{// Aggregate velocities using SGA (Eqs.~\eqref{eq:edit_proj}-\eqref{eq:V_Delta_Bar})}
   \State $z^{\text{edit}} \leftarrow z^{\text{edit}} + (t_{i-1}-t_{i})\bar{V}^{\Delta}$
   \textcolor{blue}{// Propagate ODE}
   \EndFor
   \State {\bfseries Return:} $z^{\text{edit}}$
\end{algorithmic}
\end{algorithm}

\subsection{Annealed Noise Correlation} 
\label{sec:ANC}
As discussed in Sec.~\ref{sec:roadblocks}, when setting $n_{\text{max}}=N$, the use of i.i.d.~noise leads to high-frequency jitter. We attribute this to the fact that uncorrelated noise samples in consecutive timesteps steer the process towards different edit directions. In our case, where the spatio-temporal structure of the edited video may significantly deviate from the source, this stochasticity can cause fuzziness and visible jitter. We saw in Fig.~\ref{fig:uncertainty_combined}(c) that using the same noise realization for all timesteps mitigates the high frequency jitter, but worsens the low-frequency misalignment. This is because, as discussed in Sec.~\ref{sec:SGA}, improving the low-frequency alignment requires a diverse set of noise realization to choose from. Therefore, to reduce the high-frequency jitter without worsening low-frequency misalignment, we propose an \textit{Annealed Noise Correlation} (ANC) scheduler, which introduces noise correlations that grow towards the end of the sampling process. Specifically, assuming $\tilde{w}_j$ is the $j$th noise sample at timestep $t_{i}$, then at timestep $t_{i-1}$ we set 
\begin{equation}
\label{eq:add_noise_corr}
\tilde{w}_{j}\leftarrow\sqrt{a_{t_i}}\tilde{w}_{j}+\sqrt{1-a_{t_i}}w_{j},
\end{equation}
where $\{w_{j}\}$ are i.i.d noise samples and $\{a_{t_i}\}$ is an increasing sequence such that $a_{t_N}=0$ and $a_{t_1}=1$. This guarantees that the correlation increases towards the last sampling steps, where the high frequency jitter is most prominent. The ANC module is depicted in the bottom-right part of Fig.~\ref{fig:method_pipeline}. In App.~\ref{sec:noise_corr_sched} we demonstrate the effect of the noise correlation schedule on the edit path.

\section{Experiments}
\label{sec:experiments}

\begin{figure*}[t]
    \centering
    \includegraphics[width=\textwidth]{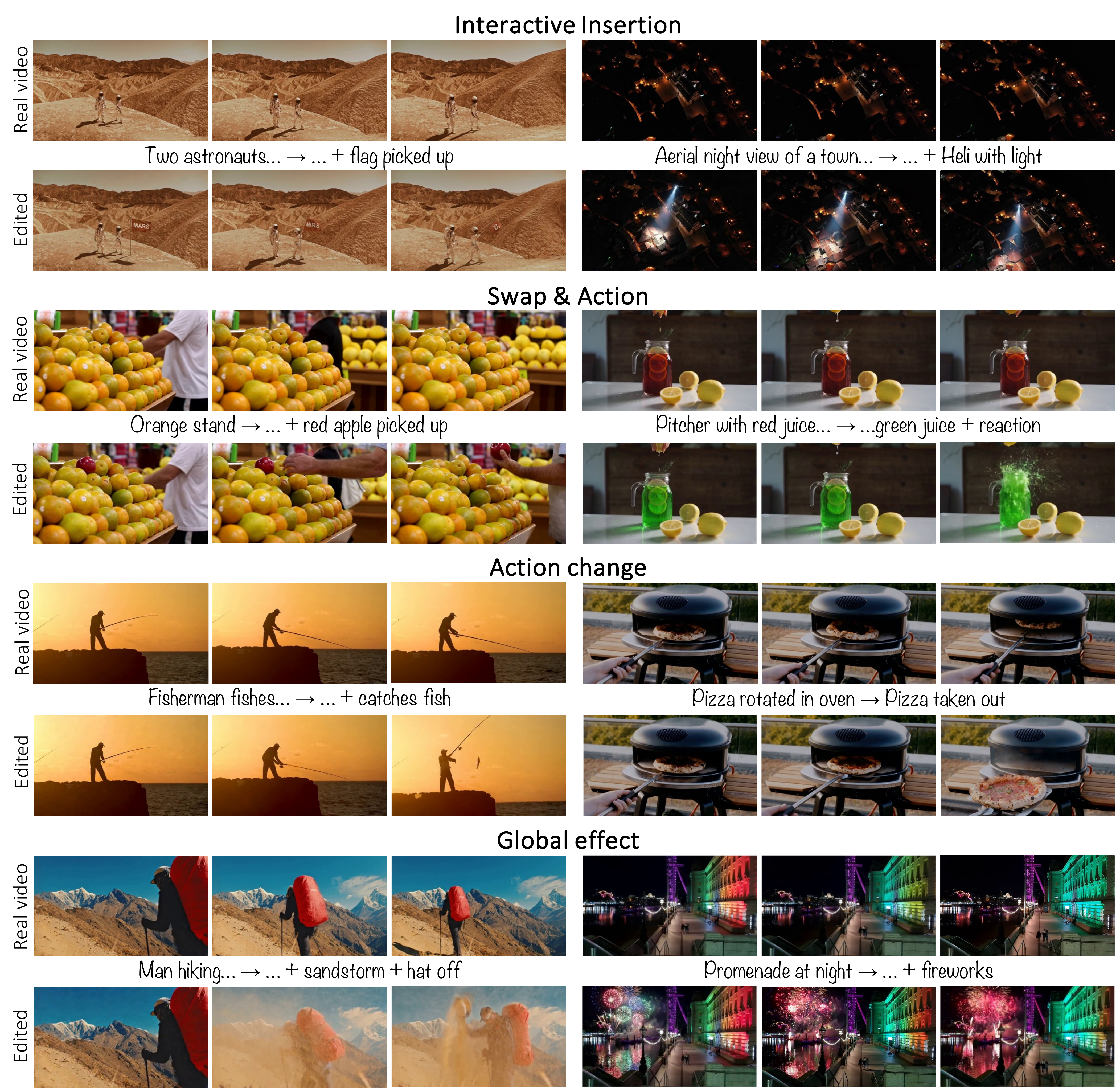}
    \caption{\textbf{\ours{} Results.} Our method supports a wide range of edits, including motion manipulation (swans), interactive object addition (horse, barrier, dinosaur), and global style changes (magma, Manga).}
    \label{fig:results}
    
\end{figure*}

\subsection{Implementation details}
\label{sec:implementation}
We conduct our experiments using the WAN2.1 14B 480p I2V model~\cite{wan2025} and provide additional qualitative results using Hunyuan 1.5 I2V~\cite{kong2024hunyuanvideo} in App.~\ref{sec:sm_hunyuan_results}. 
For the SGA module (Sec.~\ref{sec:SGA}), we use a fixed schedule of $n^{\text{SGA}}_i=5$ for the initial three steps ($i > N-3$) and $n^{\text{SGA}}_i=1$ thereafter. 
For the ANC module (Sec.~\ref{sec:ANC}), we choose $a_t$ to be linearly increasing from $0$ to $1$, starting from zero correlation at $t=1$, reaching a correlation of $1$ at $t=0.25$, and remaining constant until $t=0$.
We experiment with four hyperparameter configurations, which correspond to two options for the CFG scale parameter~\cite{ho2022classifierfreediffusionguidance} and two options for the SGA temperature $\tau$. Specifically, for the CFG parameter we use either $4.5$ and $8.5$ or $2.5$ and $4.5$ for the source and target velocities, respectively. For the temperature we use either $\tau=0.01$ or $\tau=1$. We quantitatively evaluate each of these four configurations in Sec.~\ref{sec:quant_comp} and illustrate their effects in App.~\ref{sec:sm_hyperparameter_choice}. 

\subsection{Evaluation set}

There are no existing evaluation sets for diverse video editing tasks that require significant spatio-temporal modification. We therefore curate a dataset of 71 tuples of \{source video, source text, target text, edited first frame\} in four different categories: (a)~insertion of  objects that require two-sided interaction with the original content of the video, (b)~swapping of objects with implications on the outcome of events, (c)~modifications of motion and action of objects, and (d)~global spatio-temporal effects. We manually curated the source videos from Pexels~\cite{pexels_website}, and selected target prompts to maximize diversity and ensure that each category contains at least 15 edits. We verify in App.~\ref{sec:sm_prompt_comparison} that the particular phrasings we chose for the prompts do not affect the performance of the method. The videos are 49-81 frames long, with resolution of $832\times 480$ and 16fps, as expected by the WAN model. For edits that require changes to the first frame (such as style change or object insertion), we obtain the edited first frame by querying Google's Gemini 2.5 Flash Image (Nano-Banana Pro). For the rest of the edits, the first frame is taken to be that of the source video.

\subsection{Qualitative results}
Figures~\ref{fig:teaser}, \ref{fig:results}, \ref{fig:comparisons}, and App.~Fig.~\ref{fig:sm_results} present diverse edits obtained with our method  (see videos in the SM). 
As seen, \ours{} can realistically manipulate motion (\eg making the ball enter the pocket in Fig.~\ref{fig:teaser}, bottom left), actions (\eg making the horse jump in Fig.~\ref{fig:teaser}, top left) and motion (making a fisherman catch a fish in Fig.~\ref{fig:results}, third row). 
Importantly, our method keeps the output video as similar as possible to the source video given the edit request. For example, unless specified otherwise in the text, camera motion remains similar, objects not related to the edit keep their original action, and motions follow the same patterns. We report the hyperparameters used for each of the figures in the paper in App.~\ref{sec:sm_our_hyperparam_table}.

\subsection{Competing methods}
We compare our method against FlowEdit~\cite{kulikov2025flowedit},  FlowAlign~\cite{kim2025flowaligntrajectoryregularizedinversionfreeflowbased}, editing by ODE inversion \cite{song2022denoisingdiffusionimplicitmodels}, I2V sampling, SDEdit~\cite{meng2022sdeditguidedimagesynthesis} and  Aleph~\cite{RunwayAleph2025} (a trained instruction-based video editing model). In Apps.~\ref{sec:sm_comp_dynvfx},\ref{sec:sm_comp_flowdirector} we provide additional comparisons to FlowDirector~\cite{li2025flowdirector0} and  DynVFX~\cite{yatim2025dynvfxaugmentingrealvideos}, which support only object swapping and object insertion, respectively. These are evaluated only on the relevant categories. 

For Runway Aleph, we attach the target frame as a reference image for the instruction and specify in the prompt to use it as reference for the first frame. 
Runway Aleph expects instruction-style prompts rather than source and target text captions, so we use Gemini 3 Pro to convert the pairs into an instruction prompt (see App.~\ref{sec:sm_instruction_prompts}).  For FlowEdit, FlowAlign, SDEdit, and I2V sampling, which are training-free methods, we use the same base I2V model as we use for our method. We tested all these methods with multiple hyperparameter configurations (including the reported default, where applicable) and chose the best performing option. For Runway Aleph we used the default settings in their web API. We report the parameters in App. Tab.~\ref{tab:method_comparison}.

\subsection{Qualitative comparisons}
\begin{figure}[t]
    \centering
    \includegraphics[width=\textwidth]{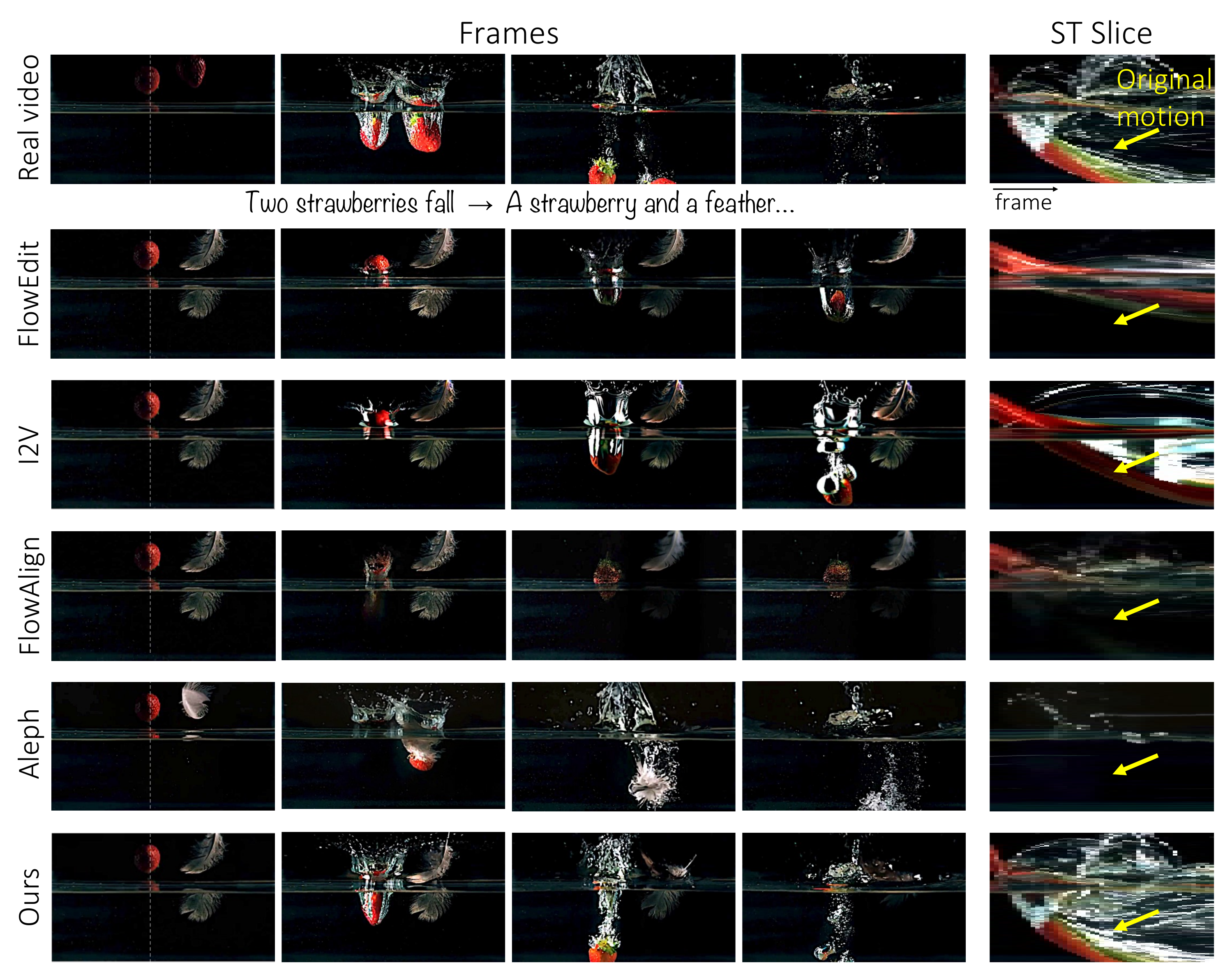}
    \caption{\textbf{Qualitative comparison.} A video depicting two strawberries falling into a water tank is edited such that the right strawberry is replaced by a feather. In contrast to competing methods, \ours{}  generates a physically plausible video, where the feather slowly descends and then floats on the water, while keeping the dynamics of the left strawberry unchanged (see App.~\ref{SM:qualitative_additional} for more comparisons).}
    \label{fig:comparisons}
\end{figure}

Figure~\ref{fig:comparisons} and App.~Fig.~\ref{SM:qualitative_additional} provide qualitative comparisons against competing baselines. 

In Fig.~\ref{fig:comparisons}, a video of two strawberries falling into water is edited to replace the right strawberry with a feather. The edit requires a significant change to the dynamics of the scene due to the different physical properties of the feather. 
As can be seen, \ours{} is the only method that manages to successfully generate a plausible edit, where the feather slowly descends until reaching the water level and then floats on the water surface. Importantly, \ours{} does not modify the dynamics of the strawberry on the left. This is in contrast to all other methods, which either cause it to disappear/fade (Aleph, FlowAlign) or change its velocity (FlowEdit, I2V). Appendix~\ref{sec:sm_additional_comparisons} provides additional examples, where the competing methods fail to generate a high-quality result.

\subsection{Quantitative comparisons}
\label{sec:quant_comp}
%\vova{add CLIP-Consistency}

\begin{figure*}
    \centering
    \includegraphics[width=\textwidth]{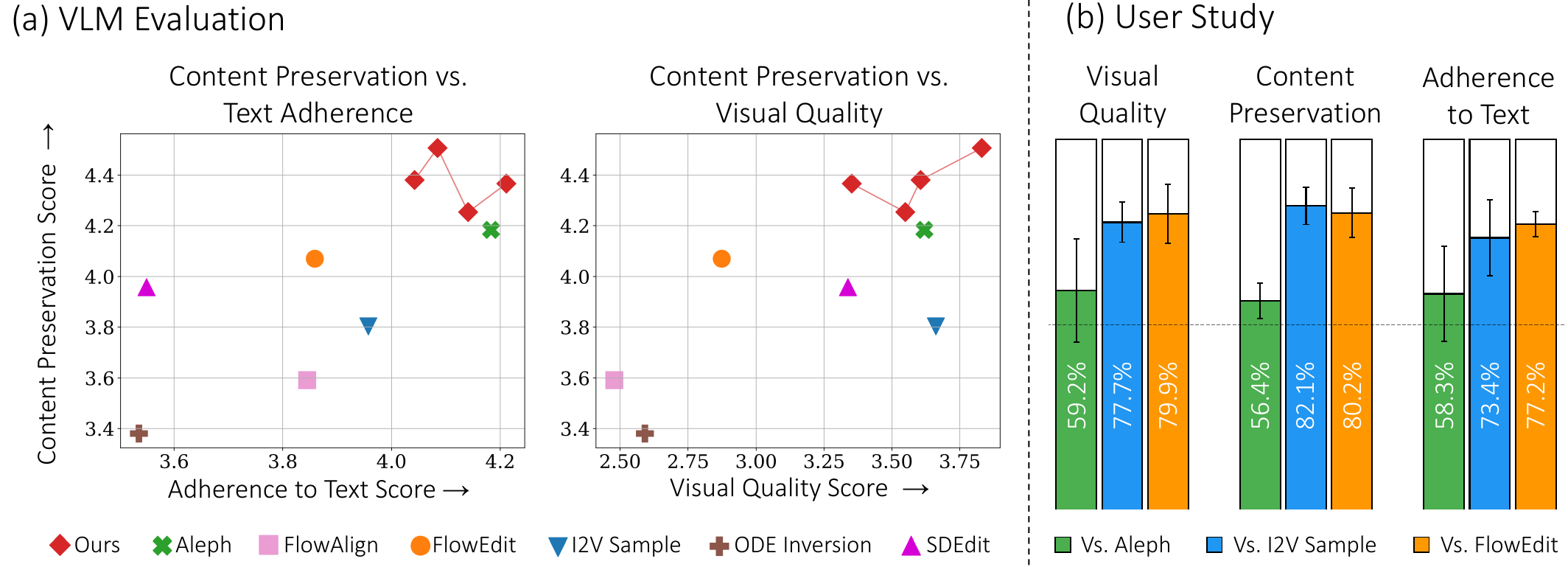}
    \caption{\textbf{Quantitative comparison.} 
    We compare \ours{} to existing methods on content preservation, text adherence and visual quality. (a) VLM ratings of these criteria show that \ours{} ranks best in content preservation, while achieving comparable text adherence and visual quality to the trained Aleph model. (b) A user study comparing \ours{} against the top three competitors shows higher preference rates for \ours{}, with the trained Aleph model being the closest competitor.
    }
    \label{fig:quantitative}
    
\end{figure*}

\paragraph{VLM-based evaluation.}
Similar to \cite{yatim2025dynvfxaugmentingrealvideos, hsu2024autovfxphysicallyrealisticvideo,garibi2025tokenverseversatilemulticonceptpersonalization}, we utilize a vision-language model (VLM) to rate the edited results with a score of 1 to 5 for each of the following three aspects: (a)~adherence to the source video (b)~adherence to the target text (c)~overall visual quality.
We use Gemini~3 Pro as our VLM and base the evaluation instructions on the ones used in \cite{peng2024dreambench}. Additional details are provided in App.~\ref{sec:sm_vlm_eval}. 
Figure~\ref{fig:quantitative} (tabular version in App. Tab.~\ref{tab:method_comparison}) shows this comparison, where our method is evaluated across all four hyperparameter configurations mentioned in Sec.~\ref{sec:implementation}. As can be seen, our method strikes the best balance between the preservation of the source video and adherence to the target text, while also achieving superior visual quality. In Fig.~\ref{fig:sm_quantitative_per_category}(a) in the appendix, we provide a breakdown of this comparison according to the four edit categories in our dataset. In all categories, our method significantly outperforms the training-free baselines, and is typically at least comparable to the trained Aleph model. 

\vspace{-0.2cm}
\paragraph{User study.}
To complement the automatic evaluation, we conducted a user study comparing our method against each of the top competing methods: Runway Aleph, FlowEdit, and I2V sampling. Here, we used a different hyperparameter configuration (among the four mentioned in Sec.~\ref{sec:implementation}) for each edit according to the extent of the required modification. For changes to larger parts of the scene we use the higher CFG configuration, and for edits requiring stronger motion changes we use the higher SGA temperature configuration (see App.~\ref{sec:sm_hyperparameter_choice}).  
In each question, the participants were presented a source video, a prompt depicting the desired edit, and two edit results, one of them being ours. The participants were asked three questions: (1)~which video best preserves the source content? (2)~which video best adheres to the target text? and (3)~which video has the best visual quality? We collected over 2400 responses from 32 unique participants. Full details on the user study can be found in App.~\ref{sec:sm_user_study_eval}. The results are reported in Fig.~\ref{fig:quantitative}, right pane. As seen, \ours{}'s results are preferred  by most users over those of the leading methods, including the trained Aleph model, in all three aspects. Per-category comparisons are provided in Fig.~\ref{fig:sm_quantitative_per_category}(b) in the appendix.

\section{Conclusion}
\label{conclusion}
We presented \ours{}, a versatile video editing framework that allows significant modification to dynamics and contents. Our method is the first training-free method that tackles this task, achieving performance that is at least comparable to the only existing trained model. 
\ours{} leverages the inversion-free paradigm, a technique that transforms the source video into its edited version through a noise-free path, but has only been explored for structure-preserving edits.
We introduced mechanisms to unlock the untapped potential of this approach for general video editing. Extensive experiments demonstrated that \ours{} achieves state-of-the-art results on complex editing tasks. 
Our method is not free of limitations, as we discuss in App.~\ref{sec:sm_limitations}. In particular, it inherits the limitations of the underlying I2V model, which often struggles with physics and leads to artifacts. Additionally, it often fails to make very large spatio-temporal modifications and simultaneously preserve regions that should not be affected by the edit. We leave improvements on these fronts to future work.

%% file: appendix.tex
\title{Appendices} 

\maketitle

% \section*{Video SM}
% We highly recommend the reader to refer to the  supplementary HTML file containing video data. It includes the full evaluation set, results and comparison videos, figures from the main text, and key figures from the appendix.

We highly recommend the reader to refer to the project's \href{https://dynaedit.github.io/}{website} containing video data, including results and comparison videos.
% \textbf{supplementary HTML file} containing video data. It includes results and comparison videos, figures from the main text, and key figures from the appendix.

\section{Additional Results}

\label{sec:sm_results_more}
\subsection{Additional results with the WAN model}

\begin{figure*}
    \centering
    \includegraphics[width=0.97\textwidth]{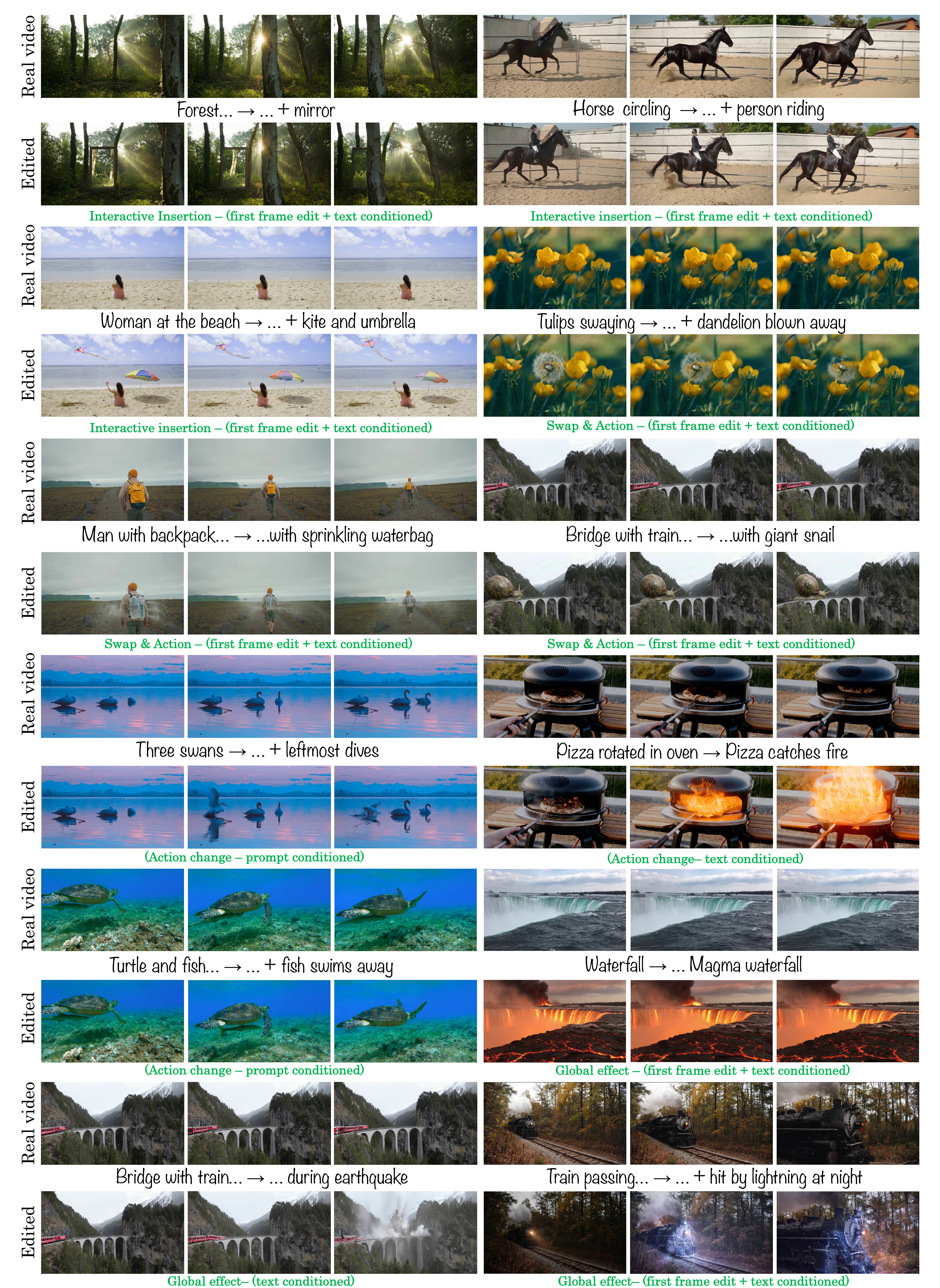}
    \caption{\textbf{Additional Results.} \ours{} is capable of diverse dynamic editing effects. For example, on the top left we insert a mirror into the forest video, showcasing reflections reactive to the scene's content. In the fourth row (left), we prompt the leftmost swan to dive, while preserving the other swans' motions. On the bottom left row, a global earthquake effect is added to the bridge scene causing the moving train to collapse, all while preserving the original camera motion.}
    % See SM HTML for the videos.}
    \label{fig:sm_results}
\end{figure*}

Figure~\ref{fig:sm_results} presents additional results obtained by our method with the WAN 2.1 14B I2V model, divided into four general editing categories: Interactive insertion, interactive swap, action change, and global effects. The videos cover diverse scenarios. 
% Please see the supplementary HTML for video results.

\subsection{Additional results with the Hunyuan model}
\label{sec:sm_hunyuan_results}
\begin{figure*}[t]
    \centering
    \includegraphics[width=\textwidth]{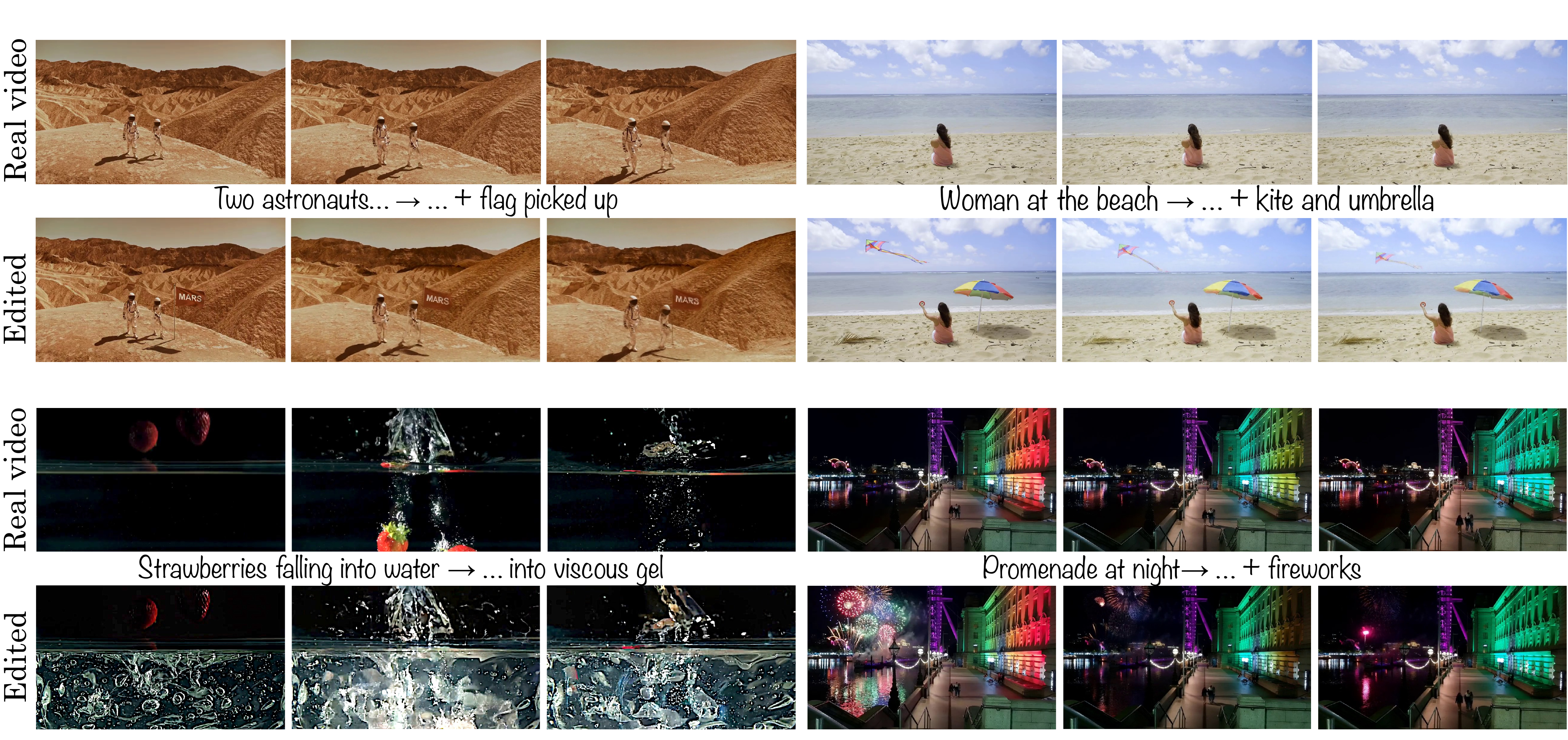}
    \caption{\textbf{Additional results with the Hunyuan I2V model.} \ours{} is model agnostic and can thus potentially leverage any I2V model. The figure illustrates diverse results obtained by \ours{} with the Hunyuan I2V model, like interactive insertion with of a Mars flag and reactive material swap (strawberries fall into gel vs.~water).}
    \label{fig:sm_hunyuan_results}
\end{figure*}

As \ours{} is model-agnostic, it can be easily adapted to work with arbitrary I2V models. To illustrate its versatility, Fig.~\ref{fig:sm_hunyuan_results} shows its use with the Hunyuan 1.5 I2V model. As can be seen, \ours{} successfully performs dynamic edits with that model, achieving good adherence to the source video as well as to the target prompt. 

\section{Additional Comparisons}
\subsection{Additional qualitative comparisons}
\label{sec:sm_additional_comparisons}
\begin{figure*}[t]
    \centering
    \includegraphics[width=1\textwidth]{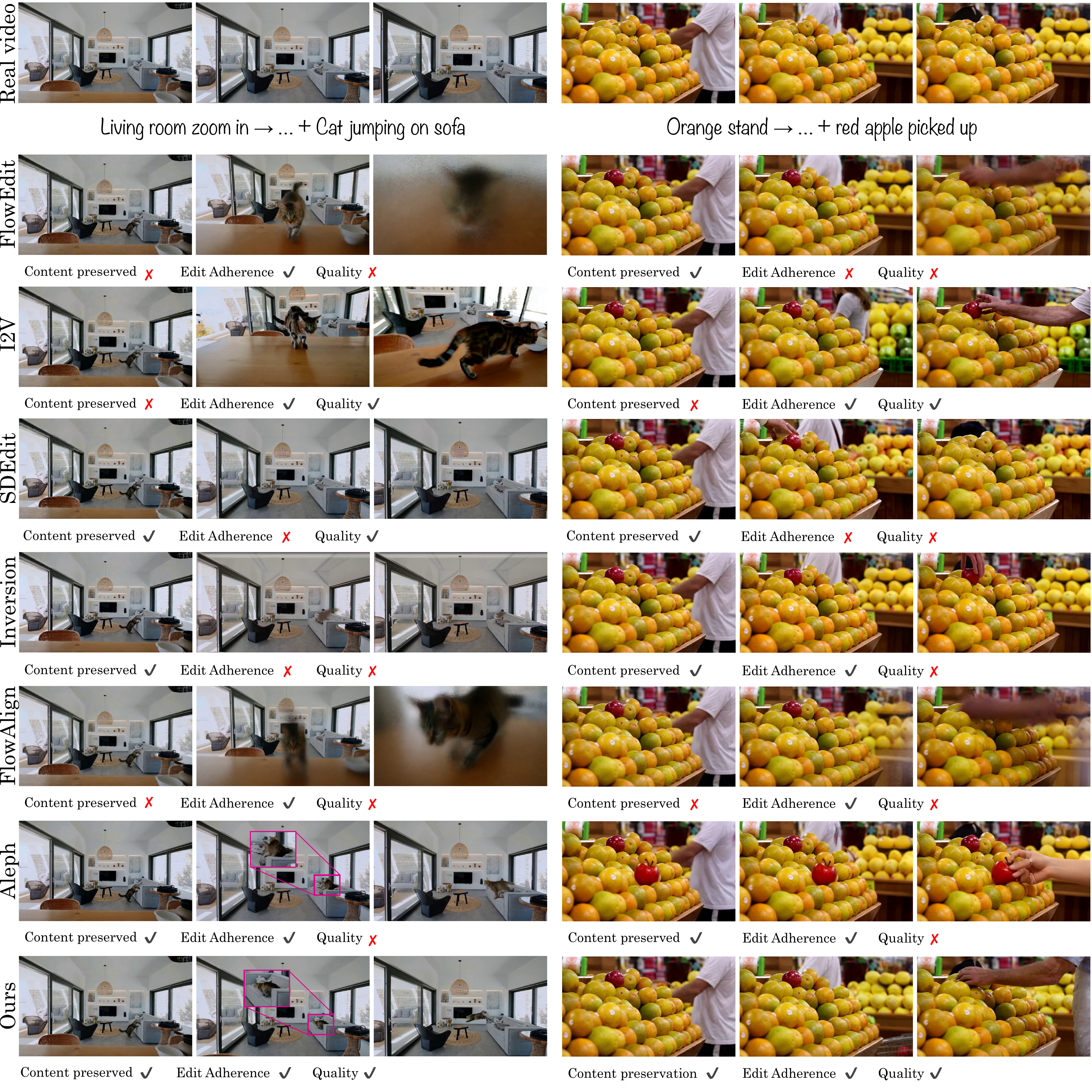}
    \caption{\textbf{Additional comparisons against the competing methods.} On the left, a cat is inserted into a zooming-in video of a living room. It is prompted to jump on the sofa. The \ours{} result is the only one to simultaneously adhere to the original camera motion, to add a cat that performs the proper action, and to exhibit no visible artifacts. On the right, an orange in a market stand is swapped into a red apple, and a person walks in to grab it. \ours{} performs the required edit without visible artifacts while preserving the background (people passing by). In both examples, the other edits do not exhibit all three properties simultaneously.}
    % See full videos and more comparisons in the SM HTML.}
    \label{SM:qualitative_additional}
    
\end{figure*}

In Fig.~\ref{SM:qualitative_additional}
 we present additional qualitative comparisons against the competing methods: FlowEdit, I2V sampling, SDEdit, Editing-by-ODE-Inversion, FlowAlign and Runway Aleph.
 % See SM HTML for the full videos and more comparisons.

\subsection{VLM evaluation - table}
\label{sec:sm_hyperparams_vlm}

In Tab.~\ref{tab:method_comparison} we present the tabular version of the quantitative VLM evaluation of Fig.~\ref{fig:quantitative}, add report several additional hyper-parameter configurations for the competing methods. The second column reports the CFG hyper-parameter configurations used for each method, and the SGA temperature parameter $\tau$ for our method (Sec.~\ref{sec:SGA}). Note that, as discussed in the main text, we employ $n_\text{max}=N$ for all inversion-free approaches, as well as for the editing-by-inversion baseline, as this configuration is crucial for enabling the spatio-temporal modifications required by the edits.

\begin{table}
\caption{Comparison vs. competing methods (with reported hyperparameters) using VLM evaluation. Best scores are highlighted in bold, second best are underlined.}
\label{tab:method_comparison}
\centering
\begin{tabular}{llccc} % Added a column for CFG
\toprule
Method & CFG $\gamma$, SGA Temp.~$\tau$ & Text Adherence & Content Pres. & Visual Quality \\
\midrule
\multirow{4}{*}{DynaEdit} & $\gamma_\text{src}=2.5, \gamma_\text{tar}=4.5,\tau= 0.01$ & 4.08 & \textbf{4.50} & \underline{3.83} \\
                          & $\gamma_\text{src}=2.5, \gamma_\text{tar}=4.5,\tau=1.0$  & 4.04 & \underline{4.38}          & 3.60 \\
                          & $\gamma_\text{src}=4.5, \gamma_\text{tar}=8.5,\tau=0.01$ & \textbf{4.21} & 4.36 & 3.35 \\
                          & $\gamma_\text{src}=4.5, \gamma_\text{tar}=8.5,\tau=1.0$  & 4.14 & 4.25          & 3.54 \\
\midrule
\multirow{3}{*}{FlowAlign} & $\gamma=18.5$            & 3.76 & 3.40          & 1.95 \\
                           & $\gamma=7.5$             & 3.84 & 3.59          & 2.47 \\
                           & $\gamma=13.5$            & 3.76 & 3.69          & 2.33 \\
\midrule
\multirow{2}{*}{FlowEdit} & $\gamma_\text{src}=2.5, \gamma_\text{tar}=4.5$   & 3.85 & 4.07          & 2.87 \\
                                   & $\gamma_\text{src}=4.5, \gamma_\text{tar}=8.5$   & 3.84 & 4.01          & 2.90 \\
\midrule
\multirow{2}{*}{I2V samp.}      & $\gamma=4.5$     & 3.95 & 3.80          & 3.66 \\
                                   & $\gamma=8.5$     & 3.95 & 3.98          & \textbf{3.88} \\
\midrule
\multirow{2}{*}{ODE Inv.}     & $\gamma_\text{src}=2.5, \gamma_\text{tar}=4.5$   & 3.53 & 3.38          & 2.59 \\
                                   & $\gamma_\text{src}=4.5, \gamma_\text{tar}=8.5$   & 3.42 & 2.80          & 1.84 \\
\midrule
Aleph                       & -        & \underline{4.18} & 4.18          & 3.61 \\
\bottomrule
\end{tabular}
\end{table}

\clearpage

\subsection{Comparison against DynVFX on object insertion}
\label{sec:sm_comp_dynvfx}
\begin{figure*}[t]
    \centering
    \includegraphics[width=\textwidth]{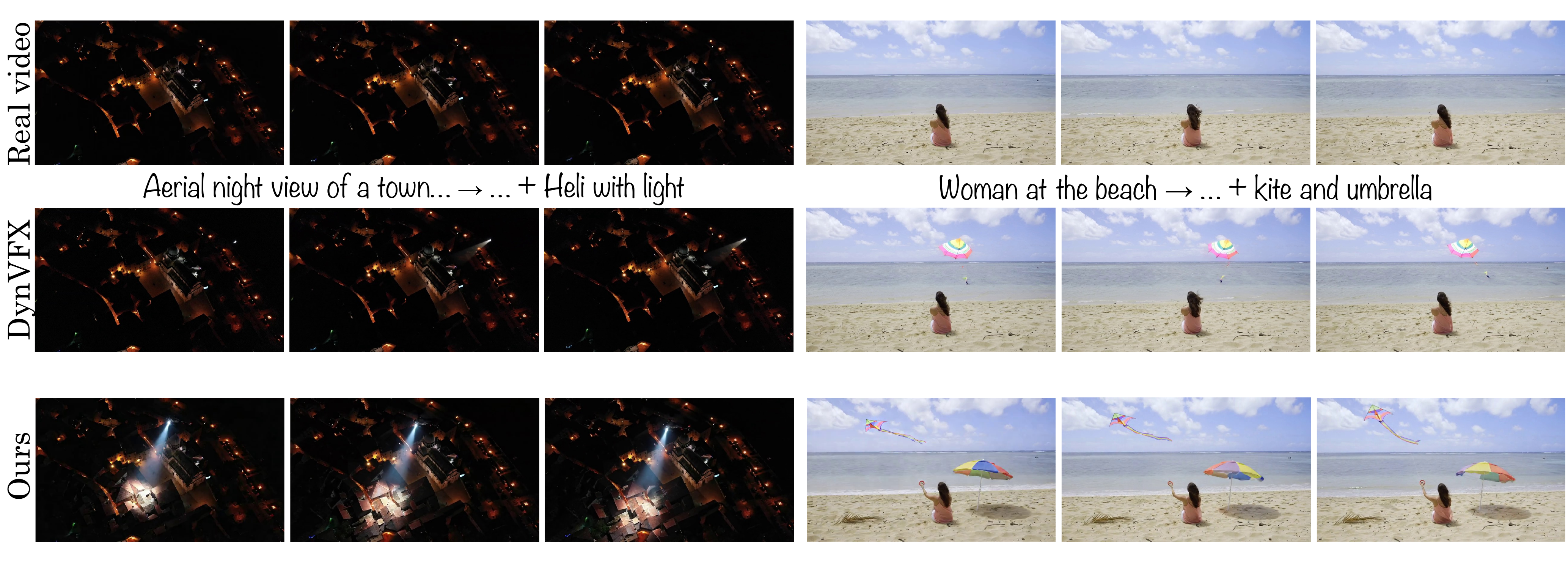}
    \caption{\textbf{Qualitative comparison against DynVFX on object insertion subset.} Our method meaninfgully integrates inserted objects into the scene. As visible by the inserted helicopter that shines light on the the city at night, or the kite and umbrella that react to the wind in the scene. This is in contrast to DynVFX, which showcases limited dynamical interactions between the inserted objects and the scene.}
    % See SM HTML for the full video comparisons.}
    \label{fig:sm_qualitative_dynvfx}
\end{figure*}

\begin{figure*}[t]
    \centering
    \includegraphics[width=\textwidth]{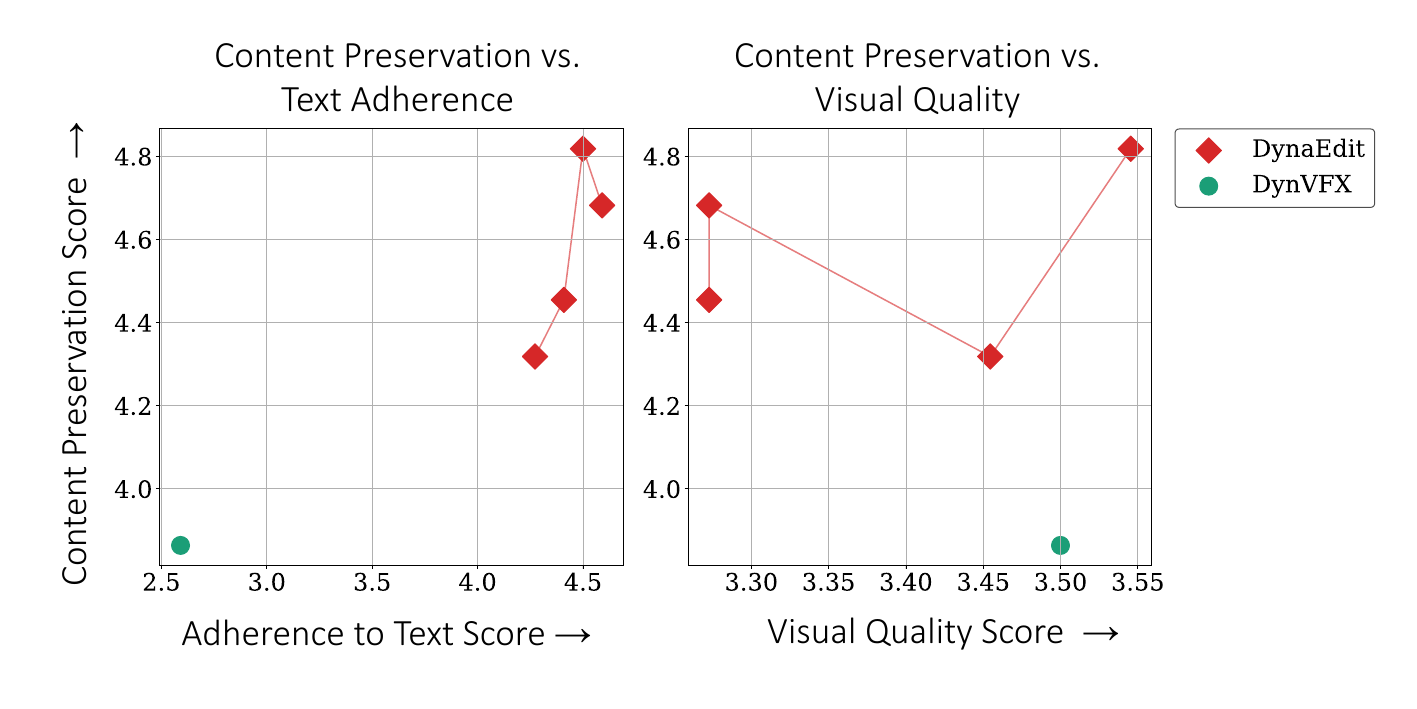}
    \caption{\textbf{Comparison against DynVFX on object insertion subset.} While DynVFX achieves a good visual quality score, the inserted objects are not able to affect the scene's outcomes. Due to this, they receive a lower text adherence score.}
    \label{fig:sm_quantitative_dynvfx}
    
\end{figure*}

Here we compare our method to DynVFX, which is an object insertion method. Since it supports only insertion based effects, we compare to it only on the relevant subset of our evaluation set (20 out of 71 videos). We use the default hyperparameters in the official implementation. Note that DynVFX is based on the CogVideoX model, which processes videos of length 49 frames at 8 fps ($\sim$6 sec.~long). The videos in our dataset contain between 49-82 frames at 16 fps ($\sim$3-5 sec.~long). Therefore, for DynVFX, we go back to the original videos, from which our clips were extracted, and re-extract 49 frames at 8 fps (this results in slightly longer versions of the videos on which \ours{} was run). 
Figure~\ref{fig:sm_qualitative_dynvfx} qualitatively demonstrates that our method is superior in terms of interactive object insertion. For instance, when inserting a helicopter that shines light on a town at night DynVFX fails to model proper light interactions, while our method does so successfully. Quantitative VLM evaluations are presented in Fig.~\ref{fig:sm_quantitative_dynvfx}. These demonstrate that our method achieves a better tradeoff between text adherence, visual quality, and content preservation.
% See more qualitative results in the SM HTML. 

\subsection{Comparison against FlowDirector on object swapping}
\label{sec:sm_comp_flowdirector}
\begin{figure*}[t]
    \centering
    \includegraphics[width=\textwidth]{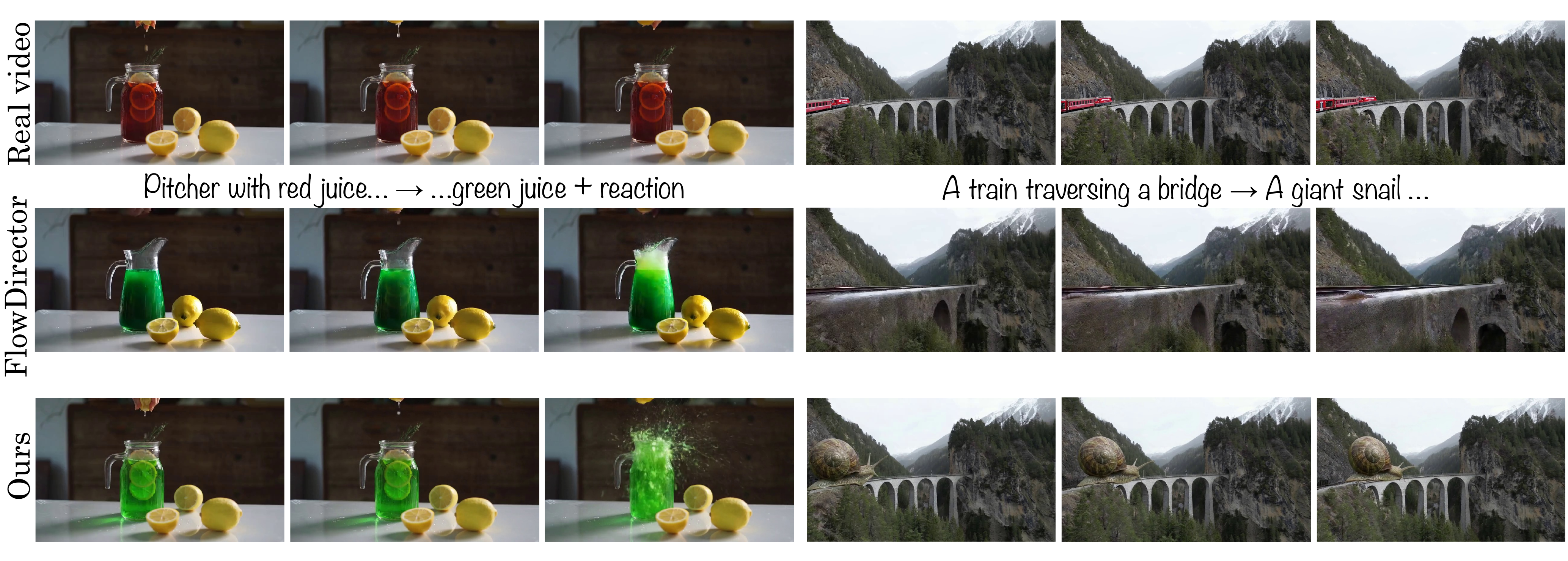}
    \caption{\textbf{Qualitative comparison against FlowDirector on object swap subset.} On the left example, the source video depicts a person squeezing a lemon into a pitcher with red juice. The edit requires changing the juice's color to green and adding a sudden chemical reaction with the lemon. Our method performs the edit successfully, depicting a realistic reaction, while FlowDirector fails to add a meaningful reaction effect due to the localized nature of it's edits. On the right, a train is edited into a giant snail. The edit prompt requires integrating an object that is not strictly bound to the source object's geometry (the train), which our method successfully follows. However, FlowDirector struggles in performing the required edit, due to the limitations imposed by it's locality.}
    % See full videos in the SM HTML.}
    \label{fig:sm_qualitative_flowdirector}
    
\end{figure*}

\begin{figure*}[t]
    \centering
    \includegraphics[width=\textwidth]{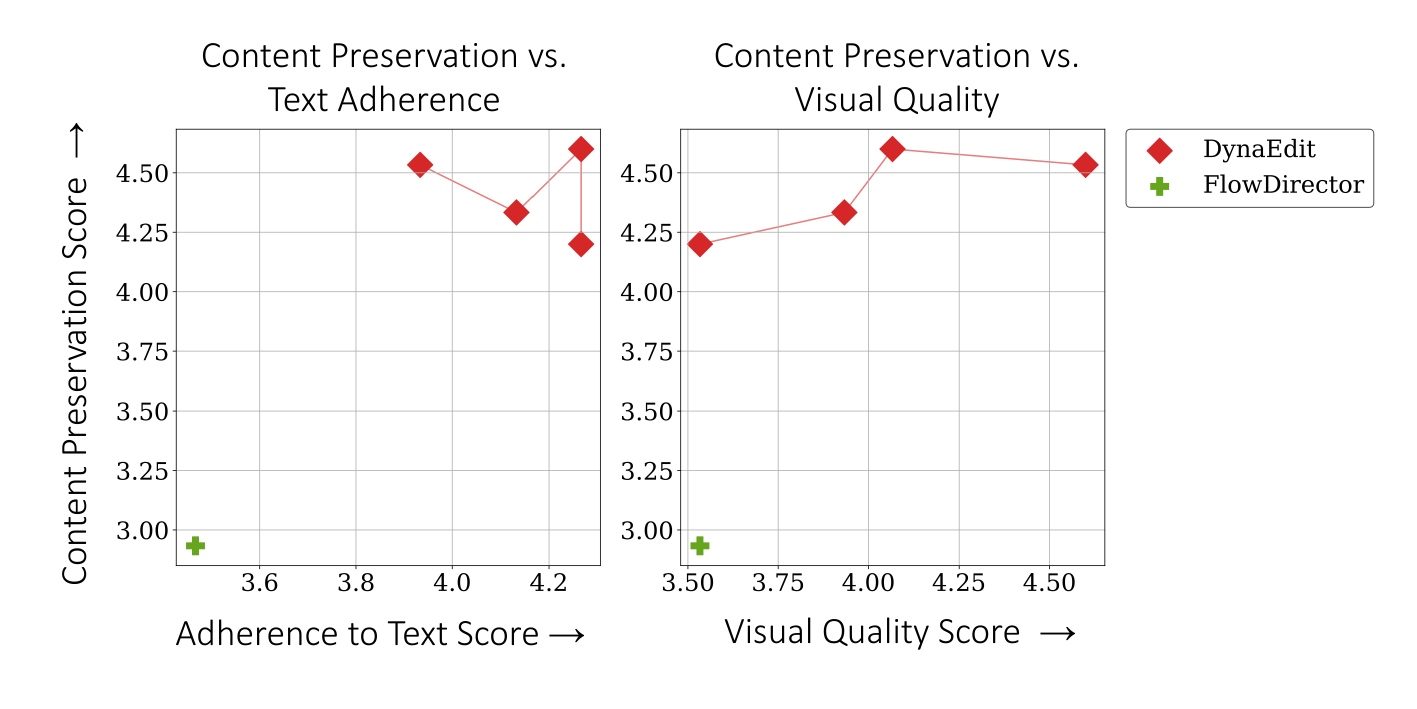}
    \caption{\textbf{Comparison against FlowDirector on object swap subset.} Due to the local nature of FlowDirector's edits, it struggles to follow prompts that require dynamic interaction with the scene, and thus generates unnatural videos. This is evident by the low text and visual quality scores. FlowDirector also performs worse in terms of content preservation, as oftentimes it struggles to maintain the swapped object's identity, or the background in close proximity to the object (see qualitative comparisons in Fig.~\ref{fig:sm_qualitative_flowdirector}).}
    \label{fig:sm_quantitative_flowdirector}
    
\end{figure*}

Here we compare against FlowDirector, which is a localized object-level editing method, using their default reported hyperparameters. We report results only on the subset of our evaluation set that includes object swap edit instructions (17 out of 71 videos). FlowDirector's editing pipeline leverages attention-based local object masking to localize the edits only to the desired object regions. While this improves fidelity to the source video, it naturally restricts the method to perform only structurally-aligned edits. As demonstrated in Fig.~\ref{fig:sm_qualitative_flowdirector}, our object swaps are interactive with the scene, while FlowDirector's edit interactions are limited to the close vicinity of the swapped object. Quantitative results are presented in Fig.~\ref{fig:sm_quantitative_flowdirector}, where we show a more favorable balance between quality, text adherence, and loyalty to the source video.
% See SM HTML for full video comparisons.

\subsection{Per-category quantitative evaluation}
\begin{figure*}[t]
    \centering
    \includegraphics[width=0.9\textwidth]{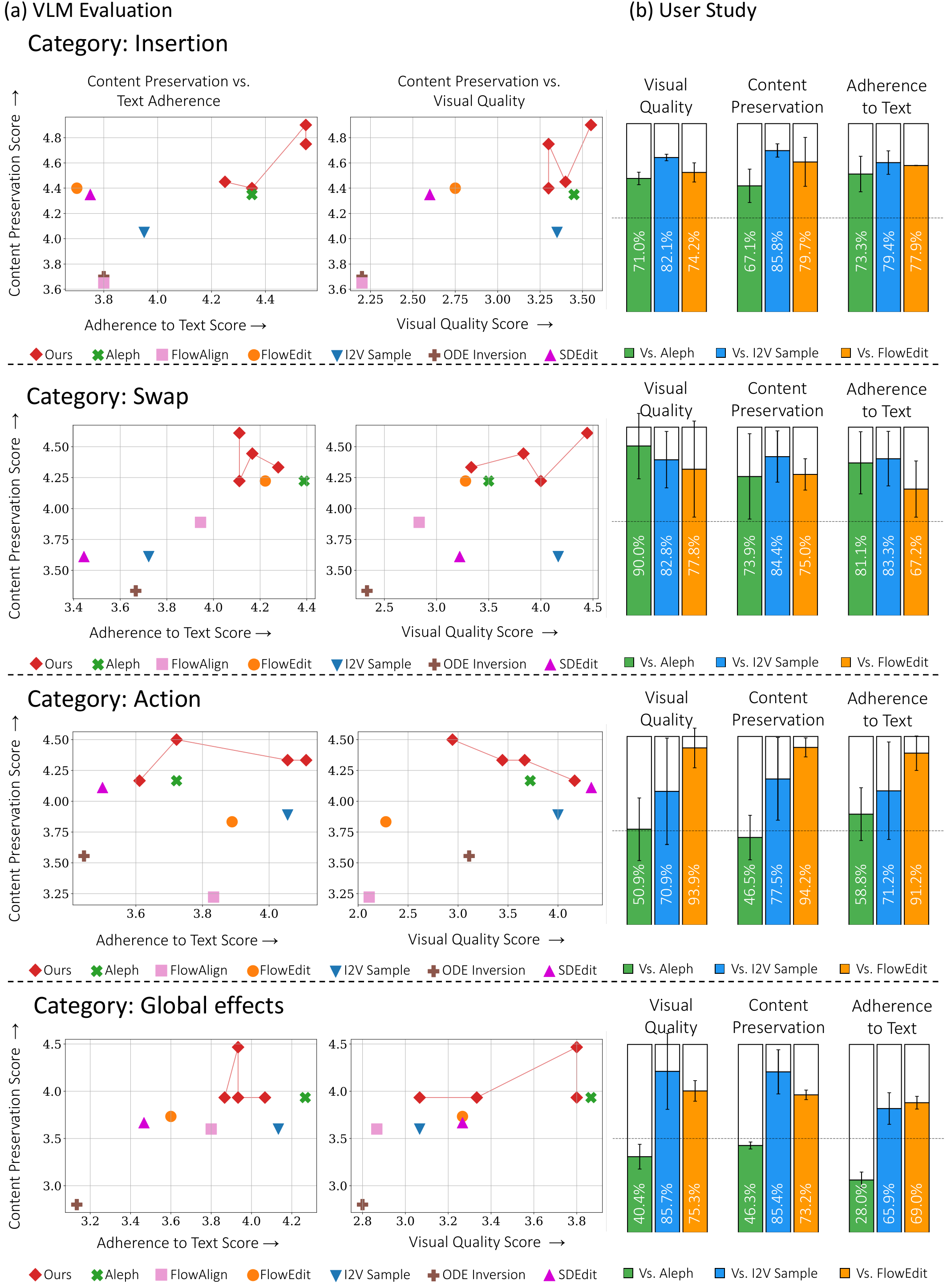}
    \caption{\textbf{VLM Evaluation and user study results per-edit-category.} (a) Per-category VLM evaluations reveal that \ours{} achieves a favorable balance between source content preservation, adherence to target text, and visual quality compared to training-free methods. Against the trained Aleph model, we achieve comparable results on most categories, with a disadvantage only in visual quality for the global effects category. (b) Per-category user study results show that on the insertion and swap categories \ours{} is favorable compared to the leading competing methods, including Aleph. For the action and global effects categories, \ours{} is preferred over the training-free approaches, and is mostly comparable to Runway Aleph, losing only in text adherence on the global effects category. }
    \label{fig:sm_quantitative_per_category}
    
\end{figure*}

In Fig.~\ref{fig:sm_quantitative_per_category} we present a per-edit-category split of the VLM and user study results that were reported in Fig.~\ref{fig:quantitative} of the main text, across the four categories in our evaluation set: Insertion, Swap, Action change, and Global effects. As seen, \ours{} outperforms the training-free methods on all categories  across the three evaluation criteria: visual quality, content preservation, and adherence to text. As for the trained Runway Aleph model, \ours{} exhibits better performance in the categories of dynamic insertion and object swap, comparable performance in the category of action change, and worse performance only in the category of global effects.

\clearpage

\section{Ablations}

\subsection{Importance of Similarity Guided Aggregation (SGA)}
\label{sec:sm_ablation_SGA}
\begin{figure*}[t]
    \centering
    \includegraphics[width=\textwidth]{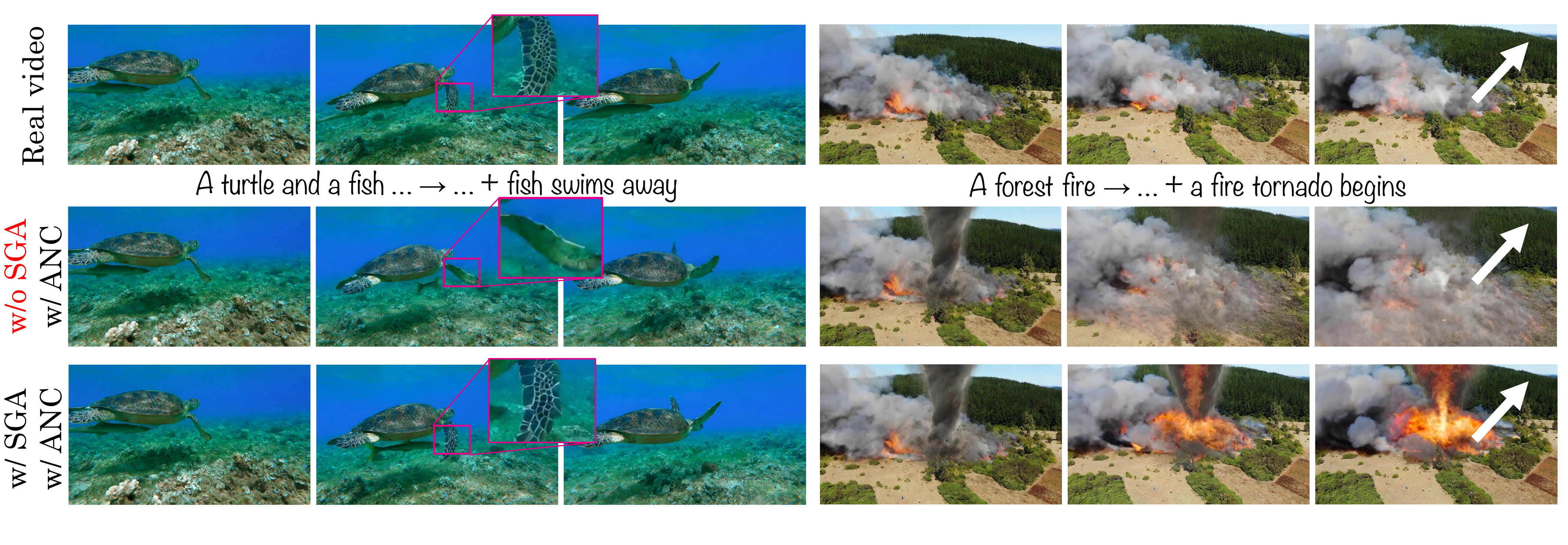}
    \caption{\textbf{Ablation of the SGA mechanism.} On the left, we edit a source video depicting synchronous motion of a sea turtle and a fish under it. The edit prompts the fish to leave the scene without changing the turtle's motion. When using the SGA mechanism, the motion of the turtle stays the same as in the source, while with regular averaging the motion changes. On the right, the edit requires adding a fire-tornado to a video of a forest fire. While strong changes are required to achieve this manipulation, the SGA mechanism keeps the original camera motion intact, as opposed to the naive averaging approach (white arrows).}
    % See SM HTML for full videos.}
    \label{fig:sm_ablation_SGA}

\end{figure*}

In Fig.~\ref{fig:sm_ablation_SGA} we qualitatively ablate the effectiveness of our SGA mechanism by comparing it to the regular averaging mechanism of FlowEdit. In both experiments the number of aggregated velocities is dictated by the same $n^\text{SGA}_i$ scheduler (as reported in Sec.~\ref{sec:implementation}). As seen, the SGA mechanism is critical for keeping good alignment with the source video in terms of both object dynamics and camera motion. 
% See SM HTML for full video results.

\subsection{Importance of Annealed Noise Correlation (ANC)}
\begin{figure*}[t]
    \centering
    \includegraphics[width=\textwidth]{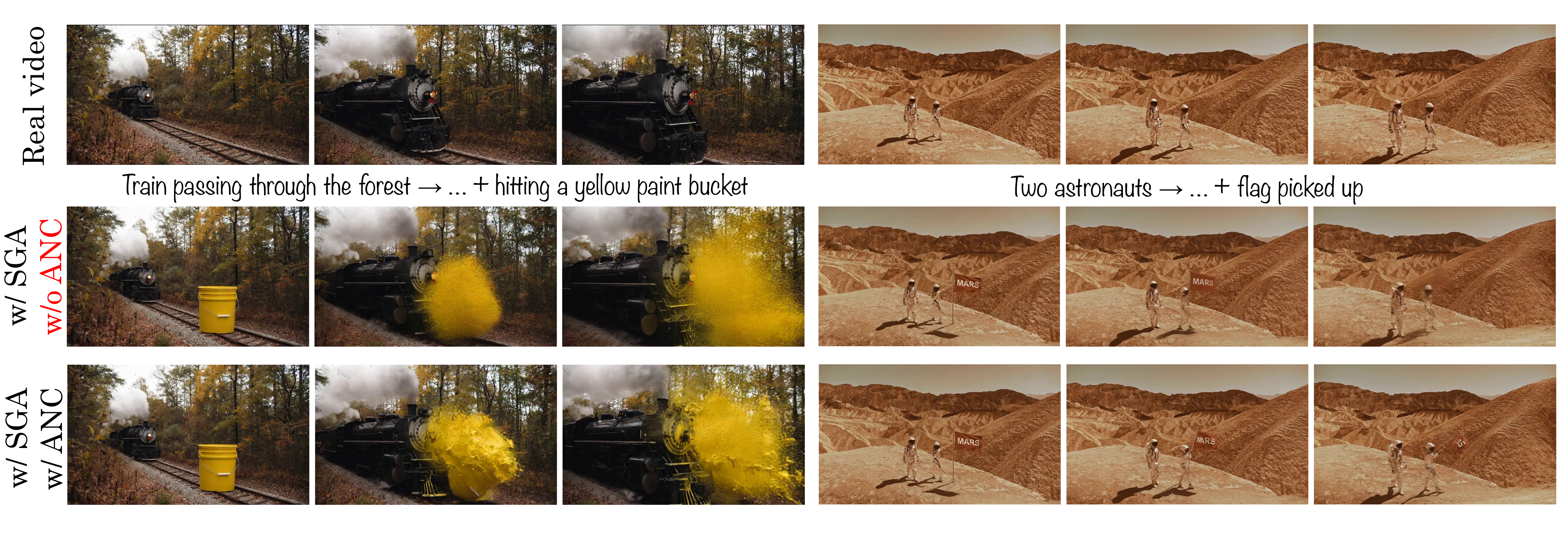}
    \caption{\textbf{Ablation of the ANC mechanism.} On the left, a source video of a train passing through a forest is manipulated by adding a yellow paint bucket to the train's path. In the i.i.d.~noise case, while the result adheres to the original motion (due to the SGA mechanism) it exhibits severe high-frequency jitter artifacts. In contrast, with ANC the edited video exhibits good fidelity in the high frequency details, evident by the visible paint particles. On the right, a flag is inserted into a video depicting astronauts walking on Mars. The edit prompt requires one astronaut to pick up the flag, while the other one to continue walking. In the i.i.d.~noise case, the flag dissolves towards the end of the video. On the other hand, our method maintains the high-fidelity dynamics of the flag throughout the whole video.}
    % See SM HTML for full videos.}
    \label{fig:sm_ablation_ANC}

\end{figure*}

In Fig.~\ref{fig:sm_ablation_ANC} we qualitatively ablate the effectiveness of ANC. We do so by comparing our full method with a version that employs i.i.d.~noise (as in FlowEdit) instead of ANC. As can be seen, the vanilla i.i.d.~scheduler leads to prominent high frequency jitter in the edited videos, whereas our ANC allows for better high-frequency fidelity.
% See SM HTML for full videos.

\begin{figure*}[t]
    \centering
    \includegraphics[width=\textwidth]{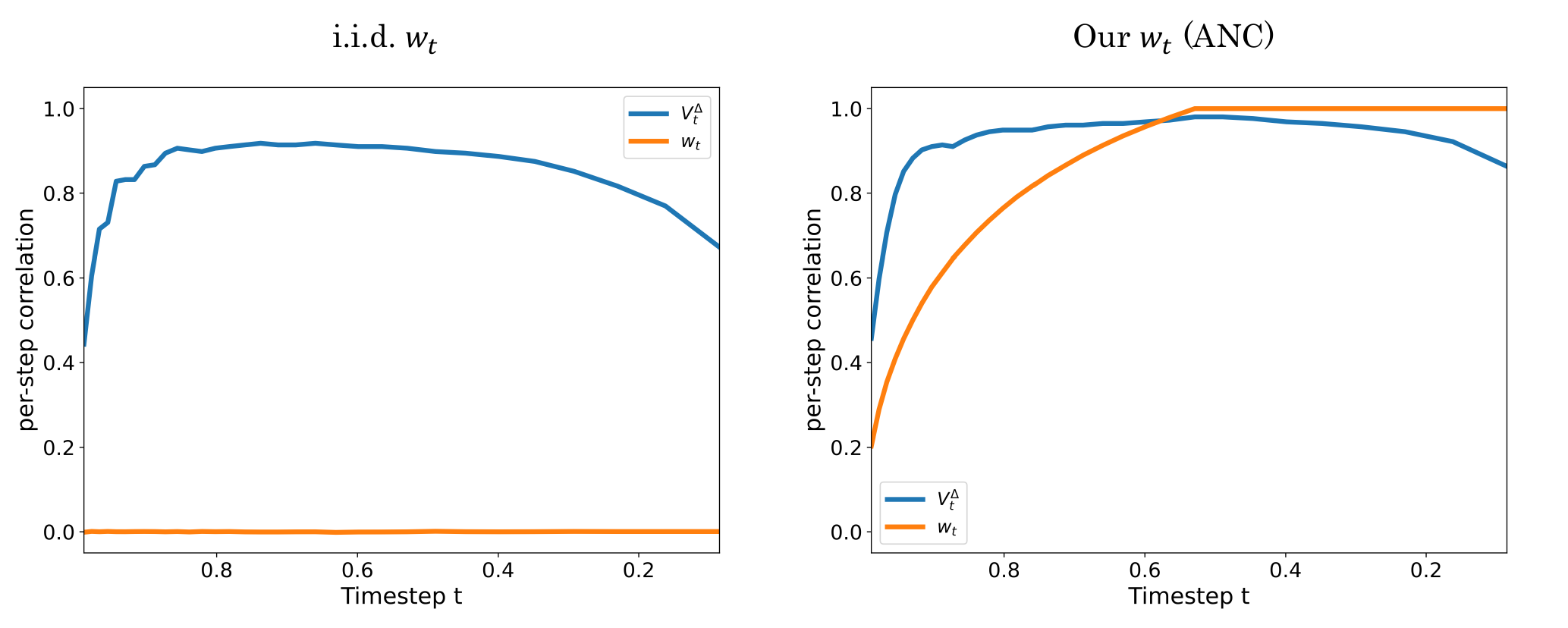}
    \caption{\textbf{Effect of noise correlations on edit velocity.} The left pane shows the effect of using i.i.d.~noise on the correlations between consecutive edit velocities. On the right, our annealed noise correlation schedule (ANC) is employed, which induces stronger correlations between consecutive noises (orange), which leads to higher correlations between consecutive edit velocities (blue) compared to the i.i.d.~case.}
    \label{sm:corr_vs_v_delta}

\end{figure*}

In Fig.~\ref{sm:corr_vs_v_delta}, we show the effect that the noise schedule has on the correlations between consecutive edit velocities $V^{\Delta}_t$. The plots show the cosine similarities between each pair of consecutive noise maps and each pair of consecutive velocity difference. As demonstrated, when our noise correlation (orange lines) is employed, the resulting velocities (blue lines) are more correlated compared to the i.i.d.~case. This behavior becomes more prominent towards the later timesteps of the generation.

\subsection{Annealed noise correlation schedules} 
\label{sec:noise_corr_sched}
\begin{figure*}[t]
    \centering
    \includegraphics[width=\textwidth]{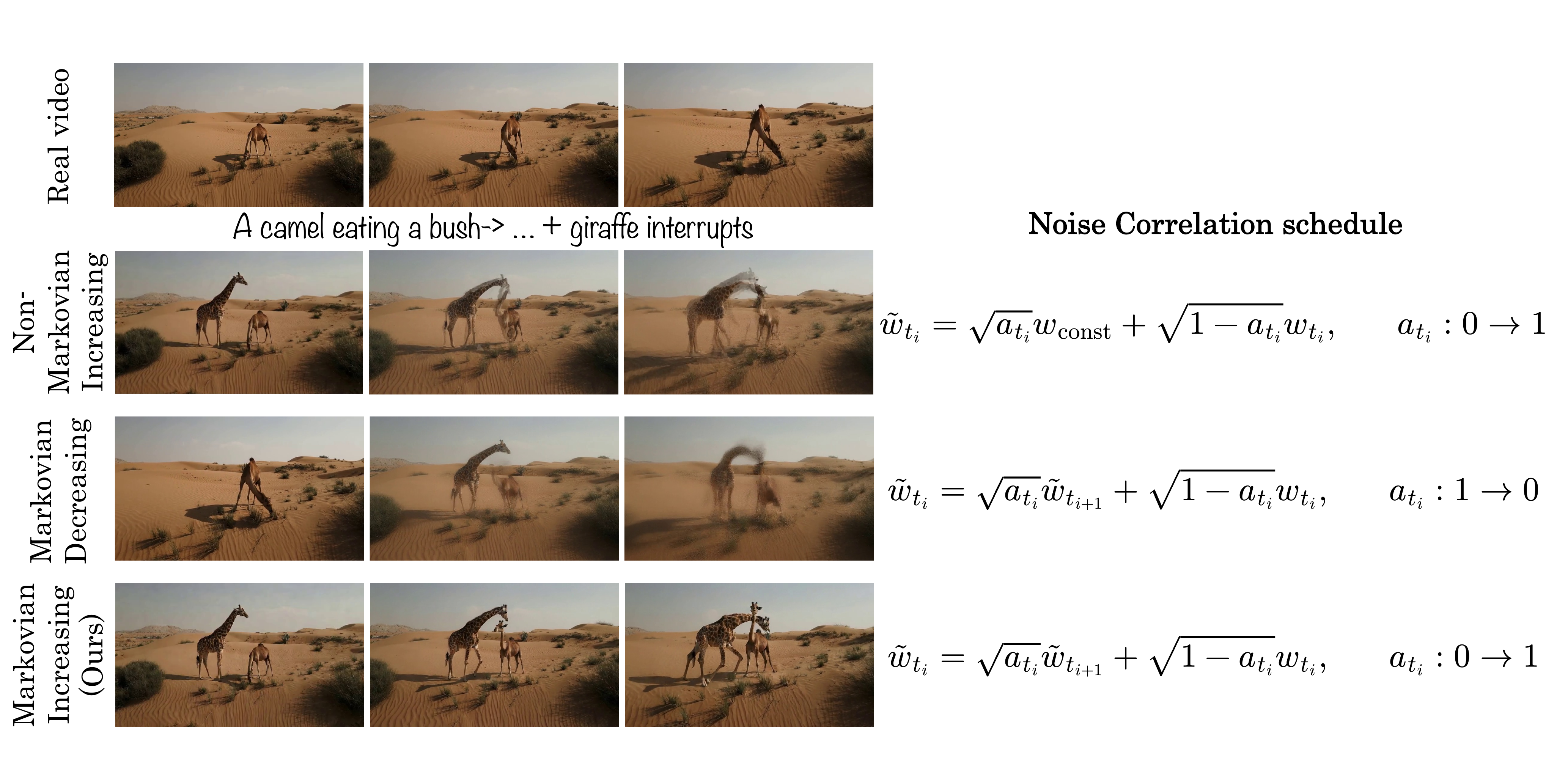}
    \caption{\textbf{Ablation of different annealing noise correlation schedules.} Three different noise correlation schedulers are compared on a video of a camel in a desert (first row). The edit prompt requires inserting a giraffe that interacts with the camel during its meal. The second row shows the result of employing non-Markovian correlations, with the same increasing coefficients $a_{t_i}$ as used in our method (see Sec.~\ref{sec:implementation}). The result showcases visible ghosting artifacts, as seen by the two giraffe necks. The third row shows the result of Markovian correlations (as in our method) but with a mirrored correlation schedule (lower correlations at later timesteps). The result exhibits high-frequency jitter artifacts. The fourth row shows Markovian correlations with an increasing schedule, which is the one employed by our method. As seen, the results are artifact- and jitter-free.}
    \label{fig:sm_corr_ablation}
\end{figure*}

In Fig.~\ref{fig:sm_corr_ablation} we ablate the effect of different noise correlation schedule choices (see right side of the figure). The first schedule is a non-Markovian increasing correlation schedule, where a different random noise map sampled at each timestep is mixed with a fixed noise map $w_\text{const}$. The correlation coefficients $a_{t_i}$ are chosen to be monotonically increasing as described in Sec.~\ref{sec:implementation}. The schedule is non-Markovian since a global map is used to correlate all steps together (to an extent dictated by $a_{t_i}$). In this case the results showcase ghosting artifacts. The second schedule is a Markovian Increasing correlation schedule, where each noise map is mixed with the previously sampled noise, but with a monotonically decreasing schedule (the exact opposite of the decreasing schedule we use in our default setting). The edited video contains jitter artifacts, showcasing the importance of introducing correlations in the later steps of the edit path, and strengthening our claim that high-frequency jitter stems from low edit correlations at later timesteps. Finally, we present the result obtained with the Markovian increasing schedule  we use in our method. As evident, the result is free of jitter or any other visual artifacts.

\subsection{SGA similarity function ablation}
\label{sec:sm_sga_sim_ablation}
\begin{figure*}[t]
    \centering
    \includegraphics[width=\textwidth]{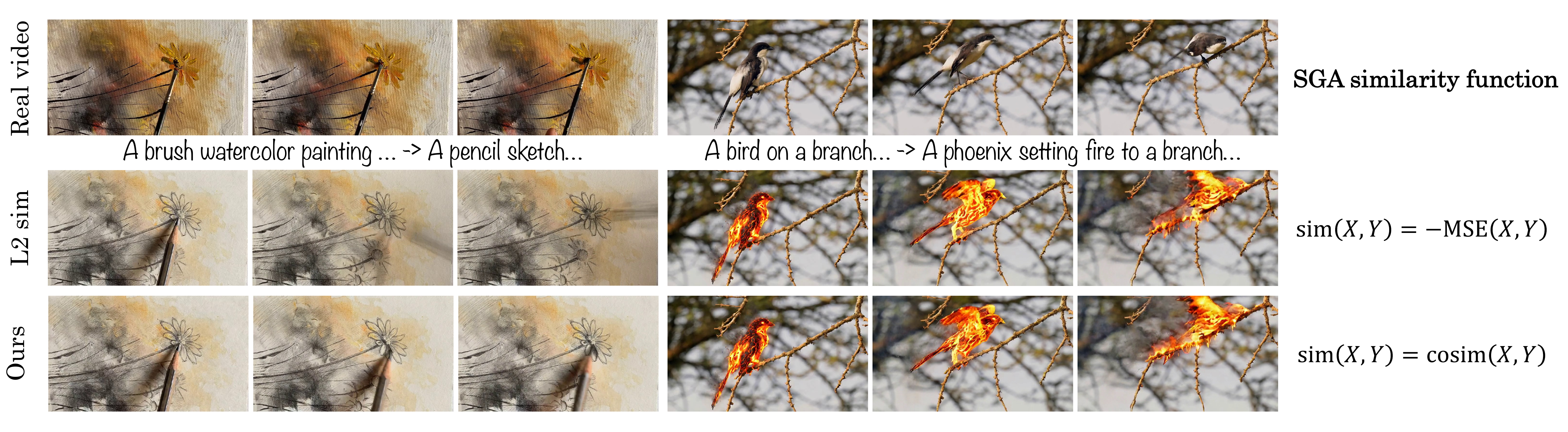}
    \caption{\textbf{Ablation of SGA similarity function choices.} Two SGA similarity functions are compared: negative MSE loss and our cosine similarity loss (see right pane). Edits are performed on a video of a paintbrush painting edited into a pencil sketch, and a video of a bird swapped into a flaming phoenix that sets fire to a branch. For the painting example, the cosine similarity function performs better, capturing the fine spatio-temporal motion of the paintbrush, as evident by the aligned motion of the pencil in the edit. This is in contrast to the MSE variant, where the pencil's location is not aligned with the paintbrush. For the phoenix example, there is little difference between the two loss functions.}
    \label{fig:sm_sga_sim_ablation}
\end{figure*}

In Fig.~\ref{fig:sm_sga_sim_ablation} we discuss the effect of different similarity functions on the SGA mechanism (Sec.~\ref{sec:SGA}). Specifically we compare the cosine similarity loss with the MSE loss (see right side of figure). As evident in the edits, sometimes the cosine similarity helps with adherence to the source when delicate objects with fine motion are present. For example, it allows preserving the paintbrush in the painting example when turning it into a pencil. However, in some cases, there is no significant difference between these loss functions. This is evident by the edit result showcasing the transformation of a bird into a phoenix.

\subsection{Robustness to edit prompt choice}
\label{sec:sm_prompt_comparison}
\begin{figure*}[t]
    \centering
    \includegraphics[width=\textwidth]{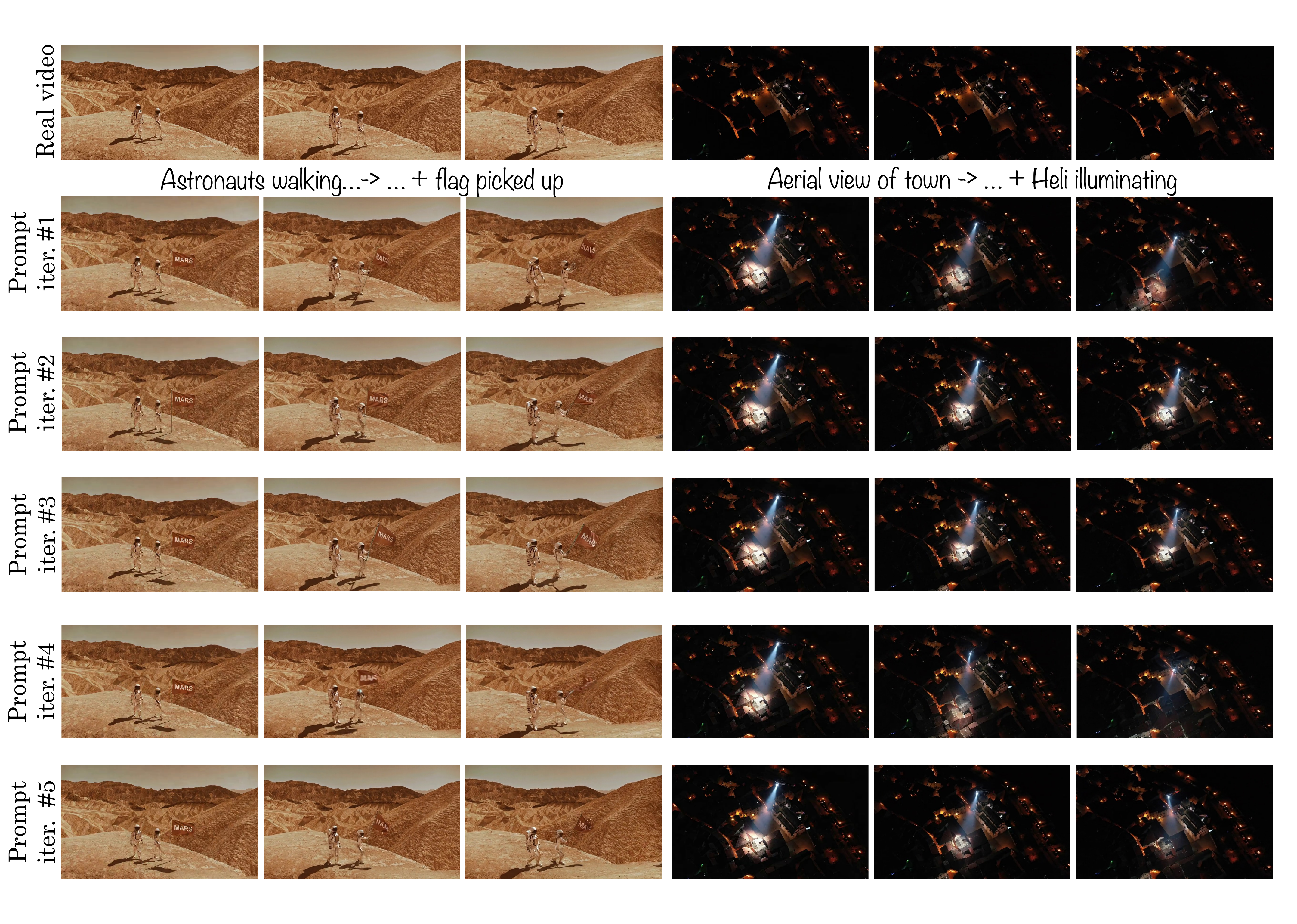}
    \caption{\textbf{Ablation of prompt robustness.} Each row depicts the editing result obtained with a different variant of the text prompts (see Sec.~\ref{sec:sm_prompt_comparison}). As can be seen, the precise phrasing of the prompt has little effect on the result.}
    \label{fig:sm_ablation_prompts}
\end{figure*}

In this section we show that the precise phrasing of the user-provided source- and target-prompts has a marginal effect on the results (as long as the phrasing conveys the same meaning). This shows that our proposed method is robust to the choice of prompt, and does not require tiresome prompt tuning. In Fig.~\ref{fig:sm_ablation_prompts} we show two different edit scenarios, with five different source-target pairs describing the same edit in different ways. As can be seen, the edit outcomes look similar regardless of the different prompting styles and lengths, strengthening our claim. We obtained these edits using Gemini~3 pro, by feeding it one source-target pair and asking it to create several different pairs with the same meaning, but with varying text styles and levels of conciseness. The source-target pairs are given in the following table.

{\scriptsize

% Define column width relative to text width (adjust 0.47 if needed)
\begin{longtable}{
    >{\RaggedRight\arraybackslash}p{0.46\linewidth}  % Left aligned, no stretching
    >{\RaggedRight\arraybackslash}p{0.46\linewidth}  % Left aligned, no stretching
}

    \caption{Prompts used for prompt robustness ablation} \label{tab:prompt_comparison} \\

    % --- Header 1 ---
    \toprule
    \textbf{Source Prompt} & \textbf{Target Prompt} \\
    \midrule
    \endfirsthead

    % --- Header 2 (Continued) ---
    \multicolumn{2}{c}{{\bfseries \tablename\ \thetable{} -- continued from previous page}} \\
    \toprule
    \textbf{Source Prompt} & \textbf{Target Prompt} \\
    \midrule
    \endhead

    % --- Footer (Continued) ---
    \midrule
    \multicolumn{2}{c}{{Continued on next page}} \\
    \bottomrule
    \endfoot

    % --- Footer (Last) ---
    \bottomrule
    \endlastfoot

    % ==========================================
    % SCENARIO 1
    % ==========================================
    \multicolumn{2}{l}{\textbf{\textit{Scenario 1: Astronauts}}} \\ % Left aligned italic header looks nicer
    \midrule
    
    Tracking shot of two astronauts traversing the Martian landscape. The terrain is a desert with mountains in the distance. & 
    Tracking shot of two astronauts traversing the Martian landscape. The terrain is a desert with mountains in the distance. The astronaut on the right hand side grabs a flag labeled 'MARS' mid-stride. The second astronaut remains oblivious to this action. \\
    
    Show two astronauts walking together on Mars. There are mountains and red sand in the background, and the camera is following behind them. & 
    Show two astronauts walking together on Mars. There are mountains and red sand in the background, and the camera is following behind them. Suddenly, the astronaut on the right spots a flag that says 'MARS' and picks it up without stopping. The other astronaut doesn't notice it happening. \\
    
    Two astronauts walking on Mars. Desert setting, mountain backdrop. Camera follows subjects. & 
    Two astronauts walking on Mars. Desert setting, mountain backdrop. Camera follows subjects. Right astronaut picks up a 'MARS' flag while walking. Left astronaut is unaware. \\
    
    A pair of space travelers stroll across the dusty red dunes of Mars, flanked by towering mountains. The viewpoint moves with them as they walk. & 
    A pair of space travelers stroll across the dusty red dunes of Mars, flanked by towering mountains. The viewpoint moves with them. The traveler on the right reaches out and snatches a flag reading 'MARS' from the ground without halting. Their companion continues walking, completely unaware. \\
    
    Generate a clip of two astronauts on Mars. Background: Mountains/Desert. Camera: Follow movement. & 
    Generate a clip of two astronauts on Mars. Background: Mountains/Desert. Camera: Follow movement. Action: The astronaut on the right must pick up a 'MARS' flag mid-walk. Condition: The other astronaut must not react to the flag. \\

    % ==========================================
    % SCENARIO 2
    % ==========================================
    \midrule
    \multicolumn{2}{l}{\textbf{\textit{Scenario 2: Aerial Night View}}} \\
    \midrule

    A bird's-eye view of a city at night, dotted with faint orange streetlights. & 
    A bird's-eye view of a city at night, dotted with faint orange streetlights. A chopper hovers overhead, projecting a powerful spotlight onto the town center that illuminates the roofs. \\
    
    Looking straight down on a dark town, illuminated only by scattered, soft orange glows. & 
    Looking straight down on a dark town, illuminated only by scattered, soft orange glows. A helicopter is present in the air, beaming a bright light down into the middle of the town, making the rooftops visible. \\
    
    An overhead perspective of a village during the night; the area is speckled with weak orange lights. & 
    An overhead perspective of a village during the night; the area is speckled with weak orange lights. There is a helicopter flying above, directing a harsh beam of light at the center of the village, lighting up the tops of the buildings. \\
    
    Top-down aerial shot of a town at night, with dim orange lamps dispersed throughout the streets. & 
    Top-down aerial shot of a town at night, with dim orange lamps dispersed throughout the streets. A helicopter flies through the sky, shining a brilliant searchlight on the town square, brightening the rooftops. \\
    
    A high-altitude view looking down at a residential area at night, where dim orange lighting is spread across the town. & 
    A high-altitude view looking down at a residential area at night, where dim orange lighting is spread across the town. A helicopter circles in the sky, casting a bright light in the middle of the town which illuminates the building roofs. \\

\end{longtable}
}

\subsection{Image-to-video conditioning}
\label{sec:sm_i2v_cond}
\begin{figure}[t]
    \centering
    \includegraphics[width=\linewidth]{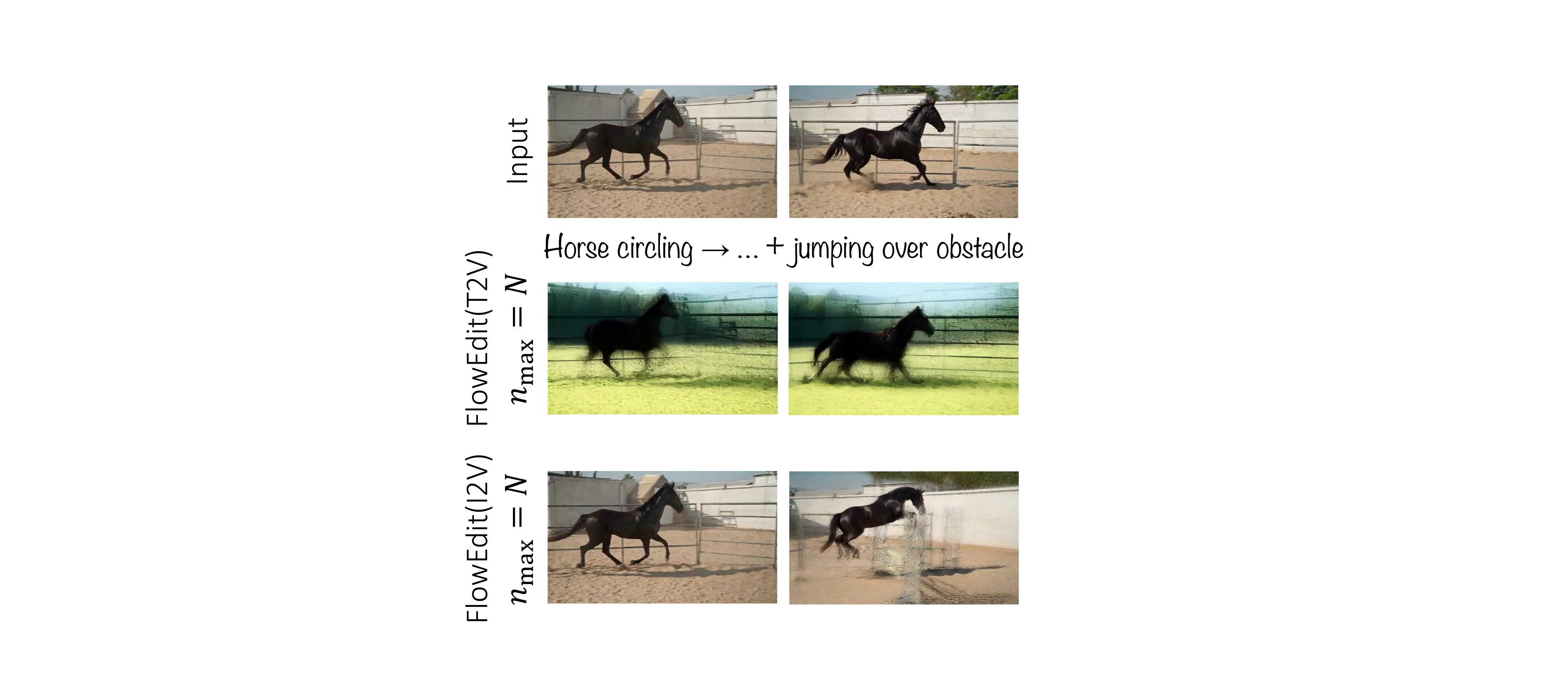}
    \caption{\textbf{Importance of I2V conditioning.} Performing strong edits ($n_{\text{max}}=N$) without an image condition for the source and target branches results in random color changes, which are evident in the edit result in the middle row. Using an Image-to-Video model conditioned on the first frame of the input video results in preserved colors, as seen in the bottom row.}
    \label{fig:SM_fe_i2v_vs_t2v}
\end{figure}

Image conditioning during video generation is a highly effective complement to the regular text conditioning, allowing more user control over the initial scene configuration, subject identities, lighting, and other things that are hard to convey with text alone. The same holds when performing editing. With I2V based editing the user can specify a target edited first frame, obtained via any image editing method that can be utilized to enforce it in the edited video. We find that this mechanism is crucial for the success of our method, where we allow edits starting from very coarse features that correspond to features like color and lighting. This can be seen in Fig.~\ref{fig:SM_fe_i2v_vs_t2v}, where we prompt a horse to jump over an obstacle instead of running in a straight line. We employ FlowEdit with the WAN2.1 T2V model, and also with the I2V model conditioned on the first frame of the source video for both source and target image conditionings. As can be seen, when visualizing intermediate edit steps $V^{\Delta}$, the T2V variant changes colors and background, while the I2V variant confines changes mainly to the horse itself without modifying colors or much of the background. This is due to the reduced edit uncertainty provided by the information in the first frame conditioning.

\section{\ours{} Hyperparameter Choices}
\subsection{Hyperparameter groups}
\label{sec:sm_hyperparameter_choice}
\begin{figure*}[t]
    \centering
    \includegraphics[width=\textwidth]{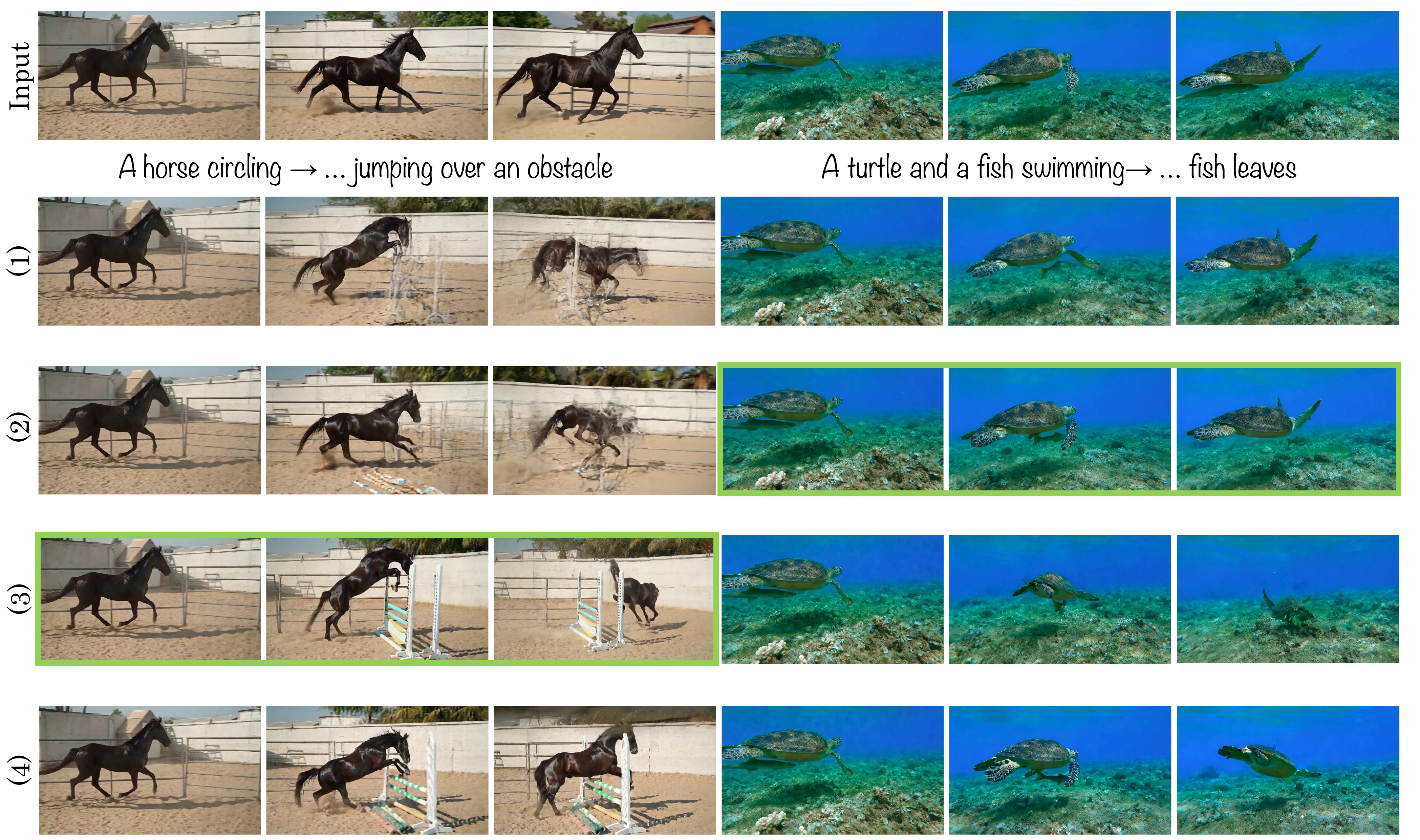}
    \caption{\textbf{Effect of Hyperparameters.} Depending on the prompt requirement,  hyperparameter configurations can sometimes be crucial to strike a favorable balance between edit adherence and loyalty to the source. On the left, the horse is prompted to jump over an obstacle. Since the horse takes up many pixels in each frame, a strong CFG is preferred (parameter sets (3),(4)). In the other cases, the edit is too weak to properly edit the horse, resulting in visual artifacts. Among parameter sets (3),(4), it can be seen that parameter set (4) (with the lower temperature) induces a strong adherence to the input video's coarse features, hindering the ability to change the horse's original trajectory. So in this case parameter set (3) (marked with {\color{green} green} frame), which uses a higher temperature, is preferable. For the turtle and fish example, where the prompt requests to change the action of the smaller fish, a weaker edit is enough to adhere to the target. So here parameter set (2) (marked with {\color{green} green} frame), which uses lower CFG and lower SGA temperature, suffices. In the other cases, the motion of the turtle changes as well. While these options are plausible in terms of text adherence, they needlessly deviate from the input video compared to the low CFG settings (parameter sets (2),(3)).}
    % See SM HTML for full videos.}
    \label{fig:sm_hyperparam_effect}
\end{figure*}

Following the discussion in Sec.~\ref{sec:implementation}, here we detail the four hyperparameter configurations we explored. In Fig.~\ref{fig:sm_hyperparam_effect} we qualitatively demonstrate the effects of these hyperparameter groups on the final edit. Importantly, we find that for the task of general text-based editing, different edit strengths lead to different plausible text-adherent outcomes, and choosing the extent of deviation from the source is for the user to decide. As reported in the main text, we propose four sets of configurations for the source and target CFG parameters (denoted here $\gamma_{\text{src}}$ and $\gamma_{\text{tar}}$) and for the SGA temperature $\tau$ on the WAN 2.1 14B I2V model: 
\begin{enumerate}
\item[(1)] $\gamma_{\text{src}}=2.5,\gamma_{\text{tar}}=4.5,\tau=1.0$: This small CFG configuration supports edits with smaller, more precise, modification requirements. The higher temperature allows more deviation from the source's coarse features (allowing more structural deviations, as well as camera motion). 

\item[(2)] $\gamma_{\text{src}}=2.5,\gamma_{\text{tar}}=4.5,\tau=0.001$: This small CFG configuration supports edits with smaller, more precise modification requirements. The lower temperature enforces stronger alignment with the source video (\eg limiting motion changes of big objects, or camera motion).

\item[(3)] $\gamma_{\text{src}}=4.5,\gamma_{\text{tar}}=8.5,\tau=1.0$: This higher CFG configuration supports edits that require larger deviation from the source video to adhere to the target prompt (such as manipulating trajectories of big objects) at the cost of weaker alignment with the input. The higher temperature allows more deviation from the source video (enabling \eg modification of camera motion).

\item[(4)] $\gamma_{\text{src}}=4.5,\gamma_{\text{tar}}=8.5,\tau=0.001$: This higher CFG configuration supports edits that require larger deviation from the source video to adhere to the target prompt (such as changing trajectories of big objects) at the cost of weaker alignment with the input. The lower temperature enforces stronger alignment with the source video (\eg limiting motion changes of big objects, or camera motion).
\end{enumerate}

For the qualitative results in the paper, as well as the user study, we use the four hyperparameter sets in the following ways.
For edits that require subtle global effects that can alter the source's coarse features (like color or motion) we found that hyperparameter set (1) is most favorable. For example, set (1) is used in the train-hit-by-lightning example in Fig.~\ref{fig:sm_results}.
For object insertion, action change, or global effects, when the objects affected are small and a strong alignment with the source video is required, we use hyperparameter set (2). An example is the insertion of the Mars flag in Fig.~\ref{fig:results}.
For edits that prompt for big action changes, or insertion of objects that strongly influence the outcome of the video, we found it useful to use hyperparameter set (3). An example is the horse-jumping-over-obstacle edit in Fig.~\ref{fig:teaser}.
For objects that affect many pixels, but do not require strong deviation from source motion, we found it beneficial to use hyperparameter set (4). An example is the fireworks edit in Fig.~\ref{fig:results}, where the people in the background preserve their motion. 
All in all, different hyperparamters could lead to plausible results on the same edit prompt, and are usually up to the user's personal preference on the tradeoff between loyalty to the source and adherence to the target text. For the user study, we used the same hyperparameter groups as for the qualitative results (see Tab.~\ref{tab:sm_hyperparam_qualitative}).

\subsection{Parameters used for the qualitative results}
\label{sec:sm_our_hyperparam_table}
In Tab.~\ref{tab:sm_hyperparam_qualitative} we report the hyperparameter groups (from Sec.~\ref{sec:sm_hyperparameter_choice}) used to obtain the qualitative results for the figures in the paper.

\begin{table}[t]
  \centering
  \caption{Hyperparameter groups for qualitative results.}
  \label{tab:sm_hyperparam_qualitative}
  \scriptsize
  \begin{tabular}{|l|c|r|} % l=left, c=center, r=right alignment
    \hline
    \textbf{Location} & \textbf{Video description} & \textbf{Parameter group} \\ \hline
    Fig.~\ref{fig:teaser}      & ``Horse jumping''      & large object, strict motion (3)      \\ \hline
    Fig.~\ref{fig:teaser}      & ``Cat playing''      & action change, large motion change (3)     \\ \hline
    Fig.~\ref{fig:teaser}      & ``Billiard ball entering''   & action change, limited motion change (4)     \\ \hline
    Fig.~\ref{fig:teaser}      & ``Nighttime beach''  & global effect, limited motion change (4)     \\ \hline
    Fig.~\ref{fig:results}      & ``Mars flag''  & small object, limited motion change (2)     \\ \hline
    Fig.~\ref{fig:results}      & ``Heli with light''  & small object, limited motion change (2)     \\ \hline
    Fig.~\ref{fig:results}      & ``Red apple picked up''  & large object, large motion change (3)     \\ \hline
    Fig.~\ref{fig:results}      & ``Juice chemical reaction''  & small object, limited motion change (2)     \\ \hline
    Fig.~\ref{fig:results}      & ``Fisherman catches fish''  & action change, limited motion change (4)     \\ \hline
    Fig.~\ref{fig:results}      & ``Pizza taken out''  & action change, limited motion change (4)     \\ \hline
    Fig.~\ref{fig:results}      & ``Hiking during sandstorm''  & global effect, large motion change (3)     \\ \hline
    Fig.~\ref{fig:results}      & ``Fireworks''  & global effect, limited motion change (4)     \\ \hline
    Fig.~\ref{fig:sm_results}      & ``Forest with mirror''  & small object, limited motion change (2)     \\ \hline
    Fig.~\ref{fig:sm_results}      & ``Woman on horse''  & small object, limited motion change (2)     \\ \hline
    Fig.~\ref{fig:sm_results}      & ``Beach with kite and umbrella''  & medium objects, limited motion change (4)     \\ \hline
    Fig.~\ref{fig:sm_results}      & ``Dandelion blown away''  & small object, limited motion change (2)     \\ \hline
    Fig.~\ref{fig:sm_results}      & ``Sprinkling waterbag''  & small object swap, limited motion change (2)     \\ \hline
    Fig.~\ref{fig:sm_results}      & ``Giant snail''  & small object swap, limited motion change (2)     \\ \hline
    Fig.~\ref{fig:sm_results}      & ``leftmost swan dives''  & subtle action change, limited motion change (2)     \\ \hline
    Fig.~\ref{fig:sm_results}      & ``Pizza catches fire''  & action change, motion change (3)     \\ \hline
    Fig.~\ref{fig:sm_results}      & ``Fish swims away''  & subtle action change, limited motion change (2)     \\ \hline
    Fig.~\ref{fig:sm_results}      & ``Magma waterfall''  & global effect,  motion change (3)     \\ \hline
    Fig.~\ref{fig:sm_results}      & ``Bridge earthquake''  & global effect, motion change (3)     \\ \hline
    Fig.~\ref{fig:sm_results}      & ``Train hit by lightning''  & subtle global effect, motion change (1)     \\ \hline
    Fig.~\ref{fig:comparisons}      & ``Strawberry and feather''  & small object, high S-T preserv. (2)     \\ \hline
    Fig.~\ref{SM:qualitative_additional}      & ``Cat jumping on sofa''  & small object, high S-T preserv. (2)     \\ \hline
    Fig.~\ref{SM:qualitative_additional}      & ``Red apple picked up''  & large object, large motion change (3)     \\ \hline
  \end{tabular}
\end{table}

\section{Creating Instruction Prompts for DynVFX and Runway Aleph}
\label{sec:sm_instruction_prompts}
As discussed in Sec.~\ref{sec:experiments}, since our evaluation set consists of source-target prompt pairs, and DynVFX and Runway Aleph require an instruction prompt, we create an analogous instruction prompt set using a VLM. Specifically, we query Gemini~3 Pro with the prompt ``Given a source prompt depicting a source video and a target prompt describing a desired edited video, analyze the differences between them, and provide an analogous instruction prompt that depicts the edit that needs to be performed.''. We then manually go over the resulting 71 instruction prompts and make sure that they faithfully convey the desired edit.
Several examples are provided in Tab.~\ref{tab:sm_conv_instr_prompt_examples}

\begin{table}[t]
  \centering
  \caption{Examples for source-target prompt to instruction prompt conversion.}
  \label{tab:sm_conv_instr_prompt_examples}
  \scriptsize
  \begin{tabular}{|p{4cm}|p{4cm}|p{4cm}|} % Fixed widths for wrapping
    \hline
    \textbf{Source prompt} & \textbf{Target prompt} & \textbf{Instruction prompt} \\ \hline
    ``Two astronauts are walking on Mars. Desert, mountains in the background. The camera follows them.''      & ``Two astronauts are walking on Mars. Desert, mountains in the background. The camera follows them. The astonaut on the right approaches a flag spelling MARS and picks it up mid walk. The other astonaut is unaware of the flag.''      & ``Make the astronaut on the right pick up a flag spelling MARS while walking. Do not change the other astronaut.''      \\ \hline
    
    ``A tracking camera shot of a forest filled with trees. The camera is moving to the right.''      & ``A tracking camera shot of a forest filled with trees. The camera is moving to the right. A big mirror is in the middle of the forest.''      & ``Place a large mirror in the middle of the forest.''      \\ \hline
    ``A scenic shot of a stone train bridge in the mountains. A red train is crossing the bridge from left to right. The camera is slowly moving forward.''      & ``A scenic shot of a stone train bridge in the mountains. A giant snail is crossing the bridge from left to right, leaving a trail of ooze. The camera is slowly moving forward.''      & ``Replace the train with a giant snail crossing the bridge and leaving a trail of ooze.''      \\ \hline
    ``A static camera shot of yellow tulips swaying in the wind. Close-up shot.''      & ``A static camera shot of yellow tulips and a dandelion swaying in the wind. Close-up shot. Suddenly a hand reaches out and picks the middle tulip, leaving the others intact.''      & ``Have a hand appear and pick the middle tulip.''      \\ \hline
    ``A static shot of a Ferris wheel in motion. The wheel is spinning, and the spokes are moving. Sky in the background.''      & ``A static shot of a Ferris wheel in motion. The wheel is spinning, and the spokes are moving. Sky in the background. Suddenly, gray clouds begin to form in the sky, turning the sky gray.''      & ``Make gray storm clouds gather in the sky.''      \\ \hline

  \end{tabular}
\end{table}

\section{VLM Evaluation Protocol}
\label{sec:sm_vlm_eval}
For the VLM evaluation reported in Sec.~\ref{sec:experiments}, we use Google's Gemini~3 Pro. 
We provide the VLM the source and edited videos and ask it to rate the edited result on a scale of 1 to 5 in three aspects. The first is adherence to the source video -- the VLM should analyze the edit requirement (what should not change based on the source-target pair or the edit instruction) and see that unnecessary changes are kept to a minimum. This includes changes in camera trajectories, motion of objects, etc. The second aspect is loyalty to the edit -- the VLM should understand what is necessary to fulfill the edit requirement, and check if the result indeed does so. This includes checking for logical interactions in case of object insertion, or logical outcomes if an action change is required, etc. The third aspect is visual quality. This includes checking for visual artifacts, like blur, flickeriness, etc. 
There is a slight difference when querying the VLM for the edit's adherence to the edit prompt between methods that accept a source-target text pair, and methods that require an edit instruction. All compared methods except for DynVFX and Runway Aleph fall into the first category. For the source-target pairs we first ask the VLM to analyze the differences between the prompts to understand the edit requirement, and then use it to evaluate the edit. For the instruction based methods, we ask the VLM to evaluate the result based on the stated edit requirement.
In the following, we give a snippet of the evaluation protocol fed to the VLM in the case of source-target (the instruction setting is similar):
\newpage

\begin{tcblisting}{
    listing only,          % Shows only the code, no "output" box
    breakable,             % Allows box to span multiple pages
    colback=gray!5!white, 
    colframe=gray!75!black, 
    title=System Prompt: Instructions,
    fonttitle=\bfseries,
    sharp corners,
    listing options={
        basicstyle=\ttfamily\scriptsize,
        breaklines=true,   % <--- THIS FIXES THE OVERFLOW
        columns=fullflexible        
    }    
}

CONTENT_PRESERVATION_INSTRUCTIONS = """
### Task Definition
You will be provided with:
SOURCE_VIDEO
SOURCE_PROMPT - a text description of SOURCE_VIDEO
TARGET_VIDEO
TARGET_PROMPT - a text description of TARGET_VIDEO

You need to carefully analyze the difference between SOURCE_PROMPT and TARGET_PROMPT to understand what should be the difference between SOURCE_VIDEO and TARGET_VIDEO.
TARGET_VIDEO should be a minimal edit of SOURCE_VIDEO to make it adhere to TARGET_PROMPT.  Elements in SOURCE_VIDEO that are not required to change to match TARGET_PROMPT should remain the same in TARGET_VIDEO.
As an experienced evaluator, your task is to evaluate how well the edited video TARGET_VIDEO is consistent with SOURCE_VIDEO, according to the scoring criteria.

### Scoring Criteria
Unless a modification is necessary to align with TARGET_PROMPT, the following criteria should be met:
1. Objects: Determine if the elements and subjects in SOURCE_VIDEO are presented in TARGET_VIDEO.
2. Alignment - the content of TARGET_VIDEO should align with the content of SOURCE_VIDEO both spatially and temporally.
3. Motions and Actions: The motions and actions in TARGET_VIDEO should be as similar to the ones in SOURCE_VIDEO as possible.
4. Camera Motion: The camera trajectory in should be identical to the one in SOURCE_VIDEO.
Do not take into consideration visual quality. Ignore visual artifacts.

### Scoring Range
Based on these criteria, a specific integer score from 1 to 5 can be assigned to determine the level of semantic consistency:
Very Poor (1): No consistency. TARGET_VIDEO completely and unneededly deviates from SOURCE_VIDEO, one could not understand from watching TARGET_VIDEO that its source was SOURCE_VIDEO.
Poor (2): Weak consistency. TARGET_VIDEO resembles SOURCE_VIDEO semantically (same types of objects, motion, etc) but the visual content is not consistent.
Fair (3): Moderate consistency. TARGET_VIDEO resembles SOURCE_VIDEO to an extent but lacks several important details or contains some inaccuracies.
Good (4): Strong consistency. TARGET_VIDEO accurately depicts most of the information from the SOURCE_VIDEO with only minor omissions or inaccuracies.
Excellent (5): Near-perfect consistency. TARGET_VIDEO remains similar to SOURCE_VIDEO with high precision and detail, with no modification other that what is necessary to adhere to TARGET_PROMPT.

### Output format
Provide only a score in the range of 1-5, and nothing else.

## Inputs
SOURCE_PROMPT: {source_prompt}
TARGET_PROMPT: {target_prompt}
"""

TEXT_ADHERENCE_INSTRUCTIONS = """
### Task Definition
You will be provided with a VIDEO, and text PROMPT describing its content.
As an experienced evaluator, your task is to evaluate the semantic consistency between VIDEO and PROMPT, according to the scoring criteria.

### Scoring Criteria
When assessing the semantic consistency between an image and its accompanying PROMPT, it is crucial to consider how well the visual content of the VIDEO aligns with the textual description in PROMPT. This evaluation can be based on several key aspects:
1. Relevance: Determine if the elements and subjects presented in the VIDEO directly relate to the core topics and concepts mentioned in the PROMPT. The VIDEO should reflect the main ideas or narratives described.
2. Accuracy: Examine the VIDEO for the presence and correctness of specific details mentioned in the PROMPT. This includes the depiction of particular objects, settings, actions, or characteristics that the PROMPT describes.
3. Completeness: Evaluate whether the VIDEO captures all the critical elements of the PROMPT. The VIDEO should not omit significant details that are necessary for the full understanding of the PROMPT's message.
4. Context: Consider the context in which the PROMPT places the subject and whether the VIDEO accurately represents this setting. This includes the portrayal of the appropriate environment, interactions, and background elements that align with the PROMPT.

### Scoring Range
Based on these criteria, a specific integer score from 1 to 5 can be assigned to determine the level of semantic consistency:
Very Poor (1): No correlation. The VIDEO does not reflect any of the key points or details of the PROMPT.
Poor (2): Weak correlation. The VIDEO addresses the PROMPT in a very general sense but misses most details and nuances.
Fair (3): Moderate correlation. The VIDEO represents the PROMPT to an extent but lacks several important details or contains some inaccuracies.
Good (4): Strong correlation. The VIDEO accurately depicts most of the information from the PROMPT with only minor omissions or inaccuracies.
Excellent (5): Near-perfect correlation. The VIDEO captures the PROMPT's content with high precision and detail, leaving out no significant information.

### Output format
Provide only a score in the range of 1-5, and nothing else.

## Inputs
PROMPT: {target_prompt}
"""

VISUAL_QUALITY_INSTRUCTIONS = """"
### Task Definition
You will be provided with a VIDEO.
As an experienced evaluator, your task is to evaluate the visual quality of the video, according to the scoring criteria.

### Scoring Criteria
1. Bluriness: Is the video sharp or blurry?
2. artifacts: Are there artifacts in the video?
3. Flickerness: Is the video temporally smooth or exhibit temporal flickering?
4. temporal consistency: Is the video temporally consistent or do objects unreasonably change in time? unnatural motions are ok but not changes to the objects.

### Scoring Range
Based on these criteria, a specific integer score from 1 to 5 can be assigned to determine the level of visual quality:
Very Poor (1): the video suffers from severe artifacts, blurriness, flickeriness, or temporal inconsistencies.
Poor (2): the video suffers from many artifacts, blurriness, flickeriness, or temporal inconsistencies.
Fair (3): the video has noticable artifacts, blurriness, flickeriness, or temporal inconsistencies but overall quality is ok.
Good (4): the video has minor artifacts, blurriness, flickeriness, or temporal inconsistencies but overall quality is good.
Excellent (5): the video has no artifacts, blurriness, flickeriness, or temporal inconsistencies. It can pass as a legit video.

### Output format
Provide only a score in the range of 1-5, and nothing else.

\end{tcblisting}

\section{User Study}
\label{sec:sm_user_study_eval}

\begin{figure}[t]
    \centering
    \includegraphics[width=0.75\linewidth]{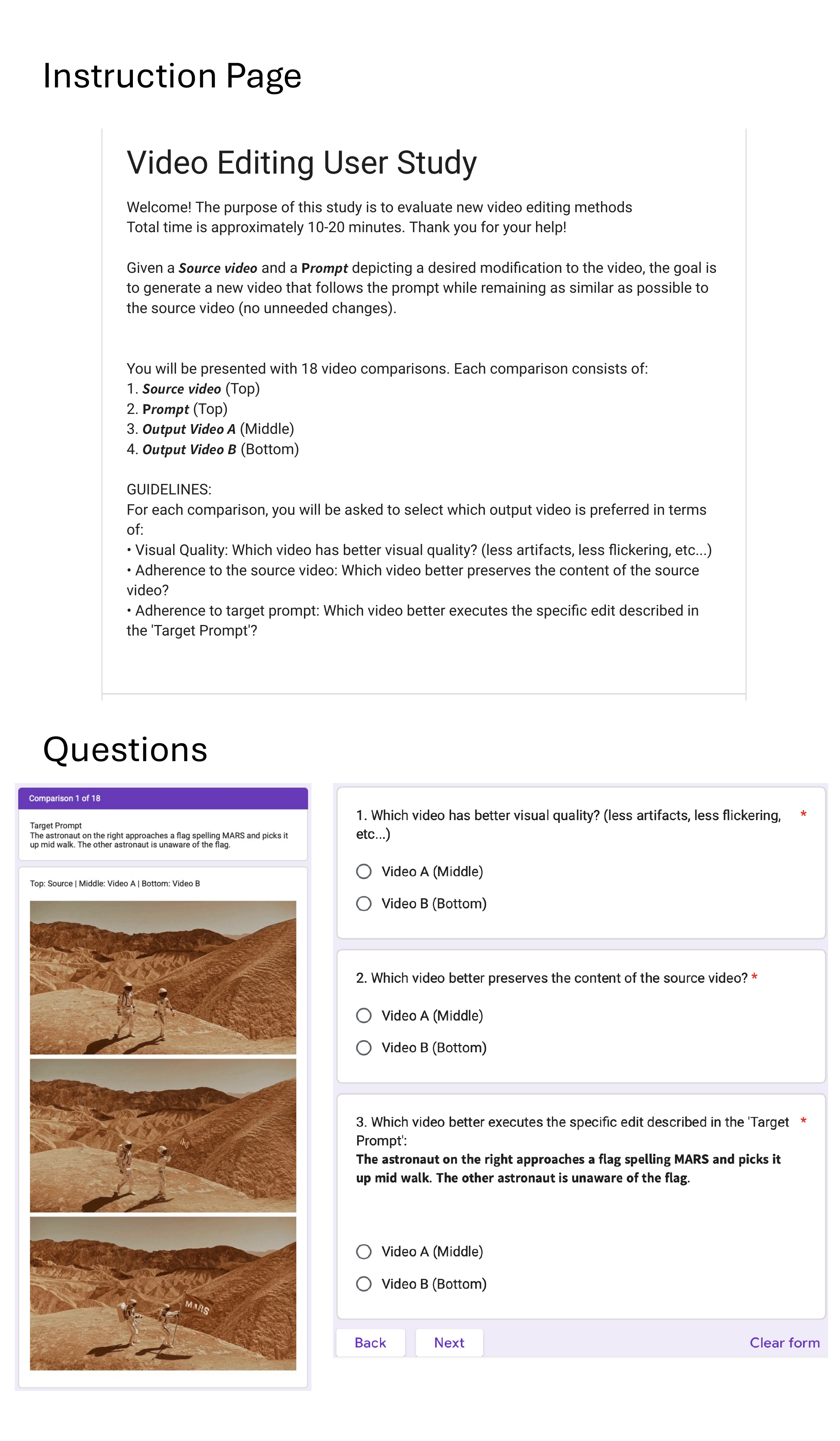}
    \caption{\textbf{User study snippet.} The first-page instruction is given at the top. The videos and following question structure are given in the bottom.}
    \label{fig:sm_user_study_snippet}
\end{figure}

As reported in the main text, to further evaluate the effectiveness of our proposed video editing method, we conducted a comprehensive user study comparing our approach against leading competing methods. The study was conducted via Google Forms. A snippet of the user study is shown in Fig.~\ref{fig:sm_user_study_snippet}.

\subsection{Study design and protocol}
The study was designed as a side-by-side comparison to assess the subjective quality of the edited videos. Participants were presented with 18 distinct video sets. Each set consisted of a \textit{Source Video} as a reference, followed by two edited results generated by our method and the baseline, labeled randomly as \textit{Method A} and \textit{Method B} to avoid bias. Importantly, the user is given an edit prompt that depicts the required difference between the source and target videos, instead of the full source-target prompt pairs. This was done to reduce the amount of redundant text that the user has to go through, and allow focusing on the edit requirement itself. We obtain the edit prompt by stripping the target prompt from text that is present in both source and target prompts. For example, for the source-target pair: \textit{source} ``Two astronauts are walking on Mars. Desert, mountains in the background. The camera follows them.'', \textit{target} ``Two astronauts are walking on Mars. Desert, mountains in the background. The camera follows them. The astonaut on the right approaches a flag spelling MARS and picks it up mid walk. The other astonaut is unaware of the flag.'', the \textit{edit prompt} would be ``The astonaut on the right approaches a flag spelling MARS and picks it up mid walk. The other astonaut is unaware of the flag.''. 
% Note that when comparing our method to Runway Aleph, which is instruction based, we also showed this edit prompt.

\subsection{Evaluation criteria}
For each comparison, participants were asked to evaluate the results based on three key dimensions:
\begin{itemize}
    \item \textbf{Visual Quality:} Assessment of technical fidelity, including the presence of artifacts, flickering, and overall realism.
    \item \textbf{Source Loyalty:} The degree to which the edited video preserves the background, structure, and original motion of the source.
    \item \textbf{Target Adherence:} The effectiveness of the method in executing the specific edit described in the provided `Target Prompt'.
\end{itemize}

\section{Limitations}
\label{sec:sm_limitations}
\begin{figure}[t]
    \centering
    \includegraphics[width=\linewidth]{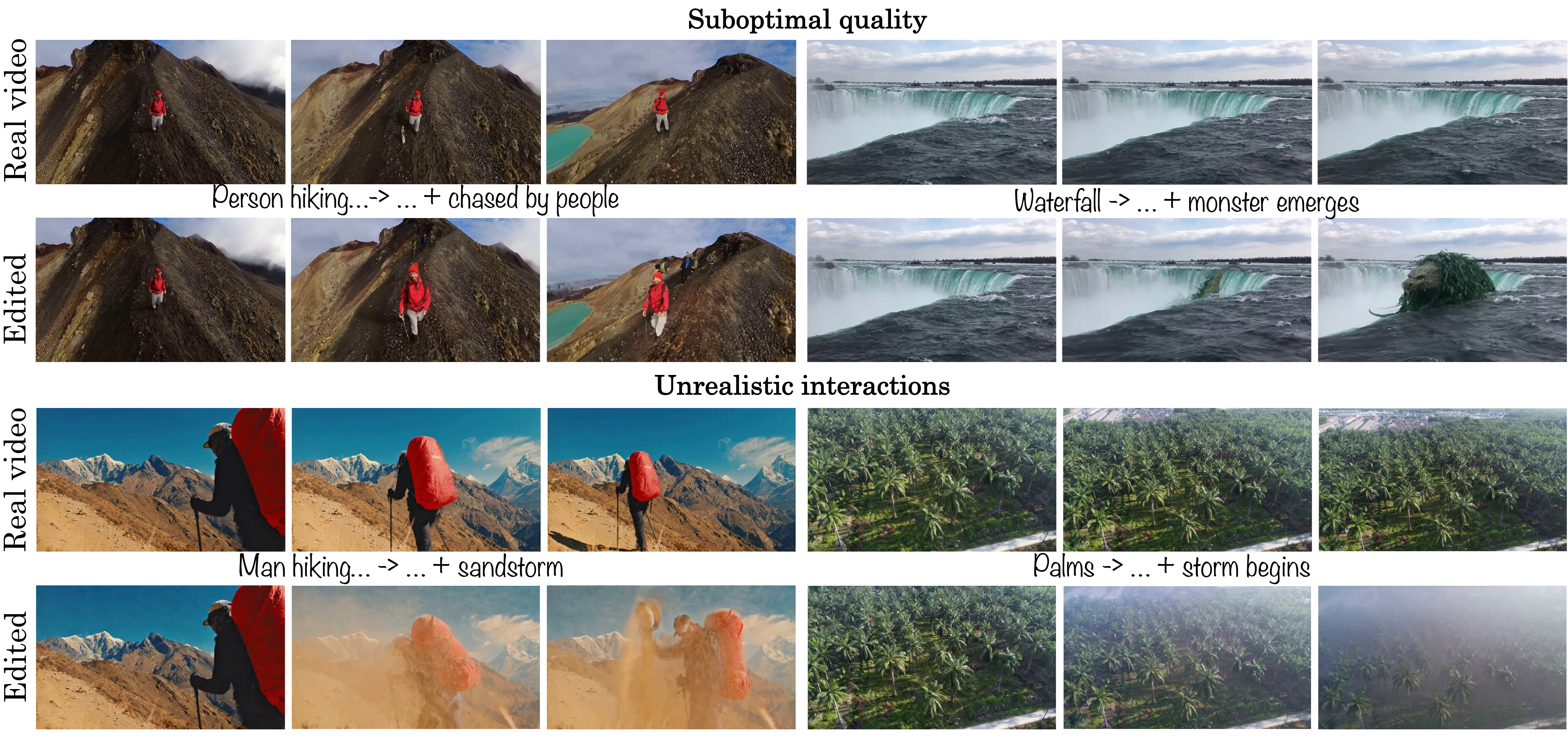}
    \caption{\textbf{Limitations arising from the base I2V model.} Top row examples showcase limited visual quality. For instance, on the right, the person's face takes up a small region in the video, so the I2V model struggles to properly generate such small details. On the top right, a video of a waterfall is edited to have a monster suddenly emerge from the water. The monster's details are not of high quality due to the model's generative capacity. The bottom examples show limitations due to the I2V model's limits in visual reasoning. For example, on the left, the man is prompted to remove his hat due to the sandstorm. After the man successfully removes the hat, another hat appears under it, which is illogical. In the bottom right, a video of palm trees is edited such that a sudden storm begins. However, the trees remain static, showcasing the model's limited understanding of wind dynamics.}
    % See SM html for full videos.}
    \label{fig:SM_limitations}
\end{figure}

Open source video models, like the WAN model we used in our experiments, have limited capacity, which sometimes causes our edits to exhibit sub-optimal visual quality or unrealistic temporal interactions. This is exemplified Fig.~\ref{fig:SM_limitations}. The top row shows results with limited visual quality. For example, on the left, the man is being chased by people with distrorted facial features, and on the right, the monster emerging from the water looks more like a giant plant. The bottom row shows examples of unrealistic interactions. For example, on the left, the man is taking his hat off during the sandstorm and reveals another hat underneath, and on the right, the palm trees remain static as in the source video despite the introduction of stormy weather. 

An additional limitation of our method is that, in some cases, achieving a favorable edit depends on the choice of hyperparameter configuration (as discussed in App.~\ref{sec:sm_hyperparameter_choice}), as these parameters control a tradeoff between source preservation, edit adherence, and sometimes visual quality. Figure~\ref{fig:sm_hyperparam_effect} shows several examples.

